\PassOptionsToPackage{dvipsnames,table}{xcolor}
\documentclass{valence}

\usepackage{amsmath, amsthm, amssymb, mathtools}
\usepackage{nicefrac}
\usepackage{bbm}
\usepackage{accents}
\usepackage{wrapfig}

\usepackage{graphicx}
\usepackage{subcaption}
\usepackage{booktabs}
\usepackage{multirow, multicol}
\usepackage{tabularx}
\usepackage{caption}
\usepackage{tikz}
\usetikzlibrary{bayesnet}
\usetikzlibrary{shapes, arrows.meta, positioning, fit, backgrounds, shadows, calc, decorations.pathreplacing}

\usepackage{algorithm}
\usepackage{algorithmic}

\usepackage{soul}
\usepackage{url}
\usepackage[textsize=tiny]{todonotes}
\usepackage{enumitem}
\everypar{\looseness=-1 }
\def\11{{\mathbf 1}}    

  \def\cG{{\mathcal G}}             \def\cD{{\mathcal D}}   \def\cP{{\mathcal P}} \def\cV{{\mathcal V}} \def\cE{{\mathcal E}}            

\def\ba{{\mathbf a}}                  \def\bs{{\mathbf s}}      \def\by{{\mathbf y}}  

                       \def\bX{{\mathbf X}} \def\bY{{\mathbf Y}} \def\bZ{{\mathbf Z}}

\def\pa{\mathrm{Pa}}

\DeclareSymbolFont{boldoperators}{OT1}{cmr}{bx}{n}
\SetSymbolFont{boldoperators}{bold}{OT1}{cmr}{bx}{n}

\newcommand{\setbrac}[1]{\left\{{#1}\right\}}
\newcommand{\circbrac}[1]{\left({#1}\right)}

\newcommand{\norm}[1]{\left\lvert #1 \right\rvert}

\theoremstyle{plain}
    \newtheorem{theorem}{Theorem}[section]

\theoremstyle{definition}

    \newtheorem{assumption}[theorem]{Assumption}

\theoremstyle{remark}

\theoremstyle{plain}
    \newtheorem{innercustomgeneric}{\customgenericname}
    \providecommand{\customgenericname}{}
    \newcommand{\newcustomtheorem}[2]{%
      \newenvironment{#1}[1]
      {%
       \renewcommand\customgenericname{#2}%
       \renewcommand\theinnercustomgeneric{##1}%
       \innercustomgeneric
      }
      {\endinnercustomgeneric}
    }

    \newcustomtheorem{customthm}{Theorem}
    \newcustomtheorem{customlemma}{Lemma}
    \newcustomtheorem{customprop}{Proposition}
    \newcustomtheorem{customcor}{Corollary}

\definecolor{geneblue}{RGB}{70, 130, 180}
\definecolor{netgreen}{RGB}{60, 179, 113}
\definecolor{llmpurple}{RGB}{147, 112, 219}
\definecolor{predorange}{RGB}{255, 140, 0}
\definecolor{evalred}{RGB}{220, 20, 60}
\definecolor{hm1}{RGB}{215, 48, 39}
\definecolor{hm2}{RGB}{252, 141, 89}
\definecolor{hm3}{RGB}{254, 224, 144}
\definecolor{hm4}{RGB}{145, 191, 219}
\definecolor{hm5}{RGB}{69, 117, 180}

\newlength\myindent
\newlength{\verticalsep}
\newlength{\horizontalsep}
\usepackage[capitalize,noabbrev]{cleveref}

\title{Knowledge Graphs and Reasoning LLMs for Finding Simple Yet Effective Transcriptomic Perturbation Predictors}

\author[1\star]{Jake Fawkes}
\author[3,4]{Liam Hodgson}
\author[2,3,4]{Jason Hartford}

\affiliation[1]{Department of Computer Science, University College London, UK}
\affiliation[2]{Department of Computer Science, University of Manchester, UK}
\affiliation[3]{Valence Labs, London, UK}
\affiliation[4]{Recursion, London, UK}

\contribution[\star]{Work done while an intern at Valence Labs.}
\correspondence{ \email{jason@valencelabs.com}}

\abstract{
   Predicting the effect of an unseen gene knockout perturbation on transcriptomic gene expression remains a highly challenging problem for virtual cell models. Recent progress has been made by leveraging biological knowledge graphs to provide a notion of similar perturbation, allowing for improved extrapolation beyond the set of training perturbations. In this work, we demonstrate that the simplest model to leverage these assumptions - a K-nearest neighbour from the knowledge graph - achieves highly competitive performance on this task, and that this can be improved further using LLMs optimised via reinforcement learning (RL) for predictive performance. Specifically, we find that the K-nearest neighbour approach beats almost all methods on out-of-distribution perturbation prediction, and when a reasoning LLM is trained via RL to make changes to the neighbourhood, it obtains equivalent performance to current state of the art methods on the cell lines from \citet{replogle2022mapping}. We also demonstrate that the RL training improves the LLM's performance on the downstream task of differential expression prediction, despite not being trained on this directly. Overall, these findings demonstrate the efficacy of knowledge graphs as model priors, and show early signs that RL can refine LLMs into generalizable tools for predicting complex biological responses.
}

\begin{document}

\maketitle

\section{Introduction}
Biology studies systems that arise from complex networks of interacting molecules, protein complexes, pathways, and cells. Much of our modern understanding of biology is built on extensive, systematic experimentation to unravel this complexity \citep{Alberts2002, Weber2012}. But these experiments are both time-consuming and expensive, so the ability to accurately predict the effects of experimental perturbations has become a key goal of computational biology \citep{lotfollahi2019scgen, lotfollahi2023cpa, Bunne2023aa, roohani2024predicting} and a necessary component in broader efforts to build \emph{virtual cell models} that attempt to recapitulate cellular behaviour \citep{bunne2024, noutahi2025virtual}. Despite these efforts, until recently modern machine learning-based methods for predicting the effects of perturbations have struggled with this task, often underperforming simple baselines such as taking the mean over all training samples \citep{wu2024perturbench,bendidi2024benchmarking, AhlmannEltze2025}.

From a causal perspective, this is unsurprising: each experimental perturbation, $i$, induces a new distribution, $P(x|\mathrm{do}(i))$, which could, in principle, change arbitrarily were it not for the constraints imposed by the biological processes that define the data-generating process.
A number of recent papers have used biological knowledge as an inductive bias, either explicitly through graph neural network (GNN) encodings of biological knowledge graphs \citep{roohani2024predicting, wenkel2025txpert} or implicitly via embeddings derived from language models \citep{wang2024modeling}. Notably, \citeauthor{wenkel2025txpert}'s GNN-based approach leverages biological knowledge graphs, such as StringDB \citep{szklarczyk2023string} and GO \citep{gene2004gene}, to attain state-of-the-art performance on the cell lines from \citet{replogle2022mapping}. But it remains unclear \emph{how} biological knowledge graphs enable generalization to these novel interventional distributions, $P(x|\mathrm{do}(i))$.

In this work, we first argue that the experimental findings of \citet{replogle2022mapping} suggest that a sparse hierarchical causal graph describes the data-generating process. The assumptions that define this data generating process imply that the average expression of each target perturbation, $i$, should be well approximated by a simple nearest-neighbour average over the local neighbourhood defined by this causal graph. 
Empirically, we find that approximating this causal graph with biological knowledge graphs yields a competitive predictor of the effects of gene perturbations---outperforming all methods except \citet{wenkel2025txpert}.

While this graph nearest-neighbour approach is appealing in its simplicity, its performance is limited by two oversimplifications: (1) biological knowledge graphs contain only a subset of the relationships between biomolecules in a cell, so the implied neighbourhoods are incomplete, and (2) because biomolecules interact in multiple ways, only a subset of genes in a local neighbourhood can be expected to behave similarly to the target perturbation. To address this, we need an approach to (1) add missing edges and (2) prune edges that should not appear in a given local neighbourhood. In principle, we can address both by optimizing the knowledge graph to maximize the accuracy of the nearest-neighbour predictor, but this involves an intractable search over $O(2^{(\text{\# genes})^2})$ possible graphs. Furthermore, it is not clear how to generalise such an approach to unseen genes.

We address this intractability by leveraging the priors that large language models (LLMs) have about genes \cite{genept}, which allow us to limit the search to candidate edges that are plausible under the LLM's prior. To do this, we train an LLM via reinforcement learning to edit gene neighbourhoods, where the model is rewarded for changes that improve predictive performance. Training via RL can then be viewed as tuning the model into a posterior over changes \citep{korbak2022rl}. We show that this training can improve the original neighbourhoods, producing a K-nearest-neighbour-style predictor that performs at a similar level to current state-of-the-art models on the cell lines from \citet{replogle2022mapping}. These results generalize to unseen perturbations.

Finally, to understand how this post-training alters the LLM, we assess it on the downstream task of PerturbQA \citep{wucontextualizing}. We find that the post-trained model shows improved performance on differential expression prediction, using the same train--test splits as in the original task to ensure evaluation on unseen genes. This adds to recent work using RL to improve LLMs' ability to reason about biology \citep{istrate2025rbio1,fallahpour2025bioreason}.
\section{Related work and Background}

{
\renewcommand{\thesubfigure}{\roman{subfigure}} 
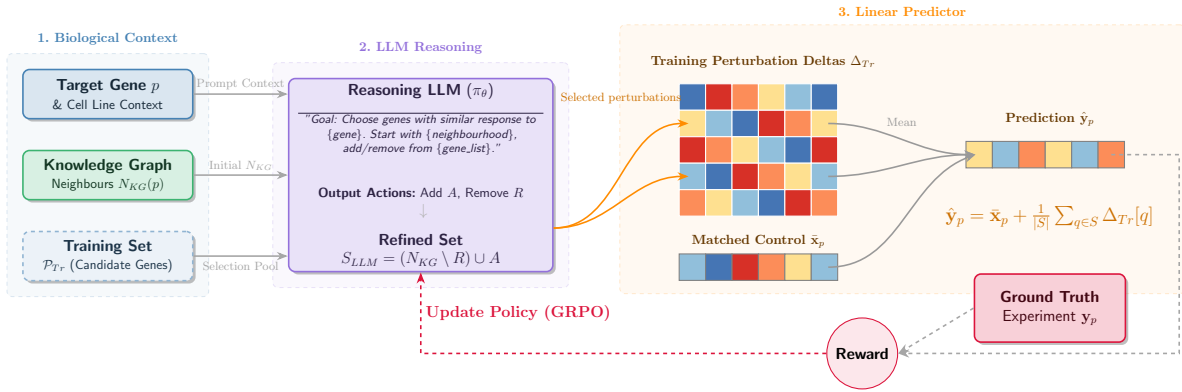
\begin{figure*}[t!]
\vspace{-0.75em}
\centering
\resizebox{\textwidth}{!}{%
\begin{tikzpicture}[
baseline=(current bounding box.north), 
    node distance=1.8cm and 2.0cm,
    node distance=1.8cm and 2.0cm,
    font=\sffamily\large, 
    >=Stealth,
    block/.style={rectangle, draw=none, rounded corners=5pt, text centered, minimum height=1.3cm, drop shadow},
    inputbox/.style={block, fill=geneblue!20, text=geneblue!30!black, draw=geneblue, very thick, minimum width=4.5cm},
    kgbox/.style={block, fill=netgreen!20, text=netgreen!30!black, draw=netgreen, very thick, minimum width=4.5cm},
    trainbox/.style={block, fill=geneblue!10, text=geneblue!40!black, draw=geneblue!60, very thick, minimum width=4.5cm, dashed},
    llmbox/.style={block, fill=llmpurple!20, text=llmpurple!30!black, draw=llmpurple, very thick, minimum width=7.0cm, minimum height=5.0cm},
    evalbox/.style={block, fill=evalred!20, text=evalred!30!black, draw=evalred, very thick, minimum width=4.0cm},
    arrow/.style={->, very thick, color=gray!70},
    dashedarrow/.style={->, very thick, dashed, color=gray!70},
    cell/.style={rectangle, draw=white, thick, minimum width=0.7cm, minimum height=0.7cm, anchor=center},
    vectorcell/.style={rectangle, draw=black!50, thick, minimum width=0.7cm, minimum height=0.7cm, anchor=center}
]

    \node[inputbox, align=center] (input) {
        \textbf{Target Gene} $p$ \\
        \normalsize \& Cell Line Context
    };

    \node[kgbox, below=0.8cm of input, align=center] (kg) {
        \textbf{Knowledge Graph}\\
        \normalsize Neighbours $N_{KG}(p)$
    };

    \node[trainbox, below=0.8cm of kg, align=center] (train) {
        \textbf{Training Set}\\
        \normalsize $\mathcal{P}_{Tr}$ (Candidate Genes)
    };

    \node[llmbox, right=2.5cm of kg, yshift=0cm, align=center] (llm) {
        \textbf{Reasoning LLM} ($\pi_\theta$)\\
        \rule{6.5cm}{0.5pt}\\
        \vspace{0.2cm}
        \small 
        \begin{minipage}{6.5cm}
        \centering
        \textit{"Goal: Choose genes with similar response to \{gene\}. Start with \{neighbourhood\}, add/remove from \{gene\_list\}."}
        \end{minipage}
        \vspace{0.4cm}\\
        \normalsize \textbf{Output Actions:} Add $A$, Remove $R$\\
        \vspace{0.2cm}
        \color{gray!60} $\downarrow$ \\
        \color{llmpurple!30!black} \textbf{Refined Set}\\
        $S_{LLM} = (N_{KG} \setminus R) \cup A$
    };
    
    \draw[arrow] (input.east) -- (llm.west |- input.east) node[midway, above, font=\small] {Prompt Context};
    \draw[arrow] (kg.east) -- (llm.west |- kg.east) node[midway, above, font=\small] {Initial $N_{KG}$};
    \draw[arrow] (train.east) -- (llm.west |- train.east) node[midway, below, font=\small] {Selection Pool};

    \coordinate (ref_top) at (input.north);
    \coordinate (ref_x) at ($(llm.east) + (3.8cm, 0)$);
    \node[below=0.5cm] at (ref_top -| ref_x) (matrix_anchor) {};
    \node[above=0.4cm of matrix_anchor, xshift=1.75cm, font=\bfseries\normalsize, color=predorange!40!black] (matlabel) {Training Perturbation Deltas $\Delta_{Tr}$};

    \begin{scope}[shift={(matrix_anchor.south west)}]
        \foreach \y [count=\yi] in {0, 0.7, 1.4, 2.1, 2.8} {
            \foreach \x [count=\xi] in {0, 0.7, 1.4, 2.1, 2.8, 3.5} {
                \pgfmathparse{int(mod(\xi+\yi*3, 5))}
                \ifcase\pgfmathresult \node[cell, fill=hm1] at (\x, -\y) {};
                \or \node[cell, fill=hm2] at (\x, -\y) {};
                \or \node[cell, fill=hm3] at (\x, -\y) {};
                \or \node[cell, fill=hm4] at (\x, -\y) {};
                \else \node[cell, fill=hm5] at (\x, -\y) {}; \fi
            }
        }
        \coordinate (row2_in) at (-0.1, -0.7); 
        \coordinate (row4_in) at (-0.1, -2.1); 
        \coordinate (row2_out) at (3.6, -0.7);
        \coordinate (row4_out) at (3.6, -2.1);
        \coordinate (matright) at (3.6, -1.4);
    \end{scope}
    
    \coordinate (llm_out) at ($(llm.east) + (0, -1.4)$); 
    \draw[arrow, color=predorange, out=0, in=180] (llm_out) to (row2_in);
    \draw[arrow, color=predorange, out=0, in=180] (llm_out) to (row4_in);
    \node[above=2.5cm, font=\small, color=predorange!80!black] at ($(llm_out)!0.5!(row4_in)$) {Selected perturbations};

    \coordinate (ctrl_pos) at ($(matrix_anchor.south west) + (0, -4.5cm)$);
    \node[above=0.4cm, font=\bfseries\normalsize, color=predorange!40!black] at ($(ctrl_pos) + (1.75, 0)$) {Matched Control $\bar{\mathbf{x}}_p$};
    
    \begin{scope}[shift={(ctrl_pos)}]
        \foreach \x [count=\xi] in {0, 0.7, 1.4, 2.1, 2.8, 3.5} {
             \pgfmathparse{int(mod(\xi+2, 5))}
             \ifcase\pgfmathresult \node[vectorcell, fill=hm1] at (\x, 0) {};
             \or \node[vectorcell, fill=hm2] at (\x, 0) {};
             \or \node[vectorcell, fill=hm3] at (\x, 0) {};
             \or \node[vectorcell, fill=hm4] at (\x, 0) {};
             \else \node[vectorcell, fill=hm5] at (\x, 0) {}; \fi
        }
        \coordinate (ctrl_right) at (3.6, 0);
    \end{scope}

    \node[right=4.0cm of matright] (vec_anchor) {}; 
    \node[above=0.4cm of vec_anchor, xshift=1.75cm, font=\bfseries\normalsize, color=predorange!40!black] {Prediction $\hat{\mathbf{y}}_p$};
    
    \begin{scope}[shift={(vec_anchor.south west)}]
        \foreach \x [count=\xi] in {0, 0.7, 1.4, 2.1, 2.8, 3.5} {
             \pgfmathparse{int(mod(\xi, 3))}
             \ifcase\pgfmathresult \node[vectorcell, fill=hm2] at (\x, 0) {};
             \or \node[vectorcell, fill=hm3] at (\x, 0) {};
             \else \node[vectorcell, fill=hm4] at (\x, 0) {}; \fi
        }
        \coordinate (vec_left) at (-0.1, 0);
        \coordinate (vec_right) at (3.5, 0);
    \end{scope}
    
    \draw[arrow, color=gray!80, out=0, in=180] (row2_out) to node[midway, above=0.2cm, font=\small] {Mean} (vec_left);
    \draw[arrow, color=gray!80, out=0, in=180] (row4_out) to (vec_left);
    \draw[arrow, color=gray!80, out=0, in=200] (ctrl_right) to (vec_left);

    \node[below=1.2cm of vec_anchor, xshift=1.75cm, font=\Large, color=predorange!80!black] (eq8) {
        $\hat{\mathbf{y}}_p = \bar{\mathbf{x}}_p + \frac{1}{|S|}\sum_{q \in S} \Delta_{Tr}[q]$
    };

    \node[evalbox, below=2.5cm of eq8, yshift=1.5cm, minimum height=1.8cm, align=center] (gt) {
        \textbf{Ground Truth}\\
        Experiment $\mathbf{y}_p$
    };

    \node[circle, fill=evalred!10, draw=evalred, thick, inner sep=5pt, left=2.0cm of gt, yshift=-1.2cm] (reward) {\textbf{Reward}};
    
    \draw[dashedarrow] (vec_right) -- ++(1.8,0) |- (reward.east); 
    \draw[dashedarrow] (gt.west) to (reward.east);
    \draw[dashedarrow, color=evalred, very thick] (reward.west) -| (llm.south) node[near end, right, font=\large\bfseries, color=evalred] {Update Policy (GRPO)};

    \begin{scope}[on background layer]
        \node[fit=(input)(kg)(train), fill=geneblue!5, rounded corners, draw=geneblue!20, dashed, inner sep=0.4cm] (bg_input) {};
        \node[above=0.1cm of bg_input, font=\bfseries\normalsize, color=geneblue] {1. Biological Context};

        \node[fit=(llm), fill=llmpurple!5, rounded corners, draw=llmpurple!20, dashed, inner sep=0.4cm] (bg_process) {};
        \node[above=0.1cm of bg_process, font=\bfseries\normalsize, color=llmpurple] {2. LLM Reasoning};

        \node[fit=(matlabel)(eq8)(matrix_anchor)(ctrl_pos), inner sep=0.7cm, fill=predorange!5, rounded corners, draw=predorange!20, dashed] (bg_pred) {};
        \node[above=0.1cm of bg_pred, font=\bfseries\normalsize, color=predorange] {3. Linear Predictor};
    \end{scope}

\end{tikzpicture}%
}
\vspace{-0.5em}
\caption{\textbf{Method Visualization} We provide LLMs with access to biological knowledge graphs and prompt them to choose genes for perturbation performance. We then optimize them via reinforcement learning using predictive performance as a reward. This finds a simple predictive model which performs at the level of state of the art models.}
\end{figure*}

\subsection{Related work }

\paragraph{Reasoning LLMs in Biology} Whilst applying reinforcement learning to LLMs for problems in AI for Biology is in its early stages, there are a few notable works in the area. Bioreason \citep{fallahpour2025bioreason} proposed a multimodal reasoning model, which takes gene embeddings from Evo2 \citep{brixi2025genome} and is trained via RL on a range of biological tasks such as disease pathway prediction. RBio \citep{istrate2025rbio1} used virtual cell models as soft verification to train a reasoning model for the task of differential expression prediction and direction of change, using the PerturbQA dataset \citep{wu2025contextualizing}. Cell-o1 trains the models to solve Cell2Sentence \citep{levine2024cell2sentence} problems, where the task is cell type classification and the input text is the names of the most expressed genes. CellForge \citep{tang2025cellforge} use a multi agent system to design dataset specific virtual cell models.

\paragraph{Perturbation Effect Prediction} 
Existing methods for predicting unseen perturbations or combinations of perturbations rely on two strategies: compositionality of embeddings / latent space representations \citep{lotfollahi2019scgen, lotfollahi2023cpa, gaudelet2024seasoncombinatorialinterventionpredictions, zuheng2024automateddiscoverypairwiseinteractions} for combinations of perturbations, and biological knowledge integration \citep{Roohani2024, Cui2024, wenkel2025txpert} or large scale pretraining on observational expressions \cite{Cui2024} or language embeddings \cite{wang2024modeling}. Like \citeauthor{Roohani2024} and \citeauthor{wenkel2025txpert} we focus on the role of knowledge graphs to assist with extrapolation, but rather than using a graph neural network, we focus on using the simplest possible model to leverage the assumption of the graph.

\paragraph{Reinforcement Learning for LLMs}
We define the LLM $\pi_\theta$ as a policy over token sequences $\mathbf{a}$. The training objective maximizes the expected reward over prompts $\mathbf{s} \sim P(\mathbf{s})$: $\mathcal{J}(\theta) = \mathbb{E}_{\mathbf{s} \sim P(\mathbf{s})} [ \mathcal{J}_{\mathrm{RL}}(\theta, \mathbf{s}) ]- \beta \mathcal{D}_{KL}[\pi_\theta \parallel \pi_{\mathrm{ref}}]$.  \emph{Group Relative Policy Optimization} (GRPO) optimizes this by sampling multiple completions, $\{\mathbf{a}_i\}_{i=1}^G$,  for each prompt $\mathbf{s} \sim P(\mathbf{s})$. It then computes the normalised advantage, $\hat{A}_i = \frac{R(\mathbf{s}, \mathbf{a}_i) - \text{mean}(R_1, \dots, R_G)}{\text{std}(R_1, \dots, R_G)}$ and uses it within the PPO loss \citep{schulman2017proximal}. Various variants have been proposed, such as dropping standard deviation normalisation and using sequence level probabilities \citep{zheng2025group}. In general these are outside the scope of this paper and more detailed summaries can be found in \citet{murphy2024reinforcement}. When training our models we use the \texttt{Verifiers} library \citep{brown_verifiers_2025} for  writing our RL environments and \texttt{Prime-RL} \citep{primeintellect2025prime-rl} for training.


\subsection{Background}


\paragraph{Transcriptomic Perturbation Prediction}
In perturbation effect prediction tasks we aim to predict properties of the distribution of gene expression profiles $P(Y | \mathrm{do}(p))$ after being experimentally perturbed by some perturbation $p\in {\cP}$. Here we focus on gene knockout perturbations, so in human cells, the size of the set of all possible perturbations $|{\cP}| = 20\,000$, from which we observe samples of gene expression profiles, $Y$, from a subset $\cP_{\mathrm{Tr}} \subset \cP$ of gene knockout experiments. We observe the associated samples, $\{P(Y | \mathrm{do}(p))\}_{p\in \cP_{\mathrm{Tr}}}$, as well as samples from unperturbed (control) cells that were collected in the same experiments $P(X | \emptyset)$. We focus on average expressions, $\mathbb{E}(Y | \mathrm{do}(p))$, so this can be thought of as aiming to complete a matrix $\bar{\bY} \in \mathbb{R}^{\norm{\cP} \times N}$ of average transcriptomic expressions where we only observe a subset of rows in this matrix corresponding to the knockouts in $\cP_{\mathrm{Tr}}$.
The controls can be thought of as a second matrix $\bar{\bX} \in \mathbb{R}^{\norm{\cP} \times N}$ where each row corresponds to the average transcriptomic expression for the paired controls for a perturbation $p$, denoted $\bar{x}_p$. We get to observe all entries for this matrix. 
In practice, biologists are usually most interested in a perturbed gene's \emph{differential expression}, which is the difference between its perturbed expression profile and that of its matched controls. We refer to this as the perturbation delta and denote $\bar{\delta}_p \coloneqq \bar{y}_p- \bar{x}_p$. 

To quantify the agreement between true and predicted perturbation deltas we use Pearson delta and Pearson delta based retrieval as our main metrics, following the work of \citet{wenkel2025txpert}. Pearson delta is defined as the Pearson correlation coefficient between our predicted delta and the true delta, denoted by $\Delta(\hat{\delta}_p,{\delta}_p)$. The retrieval metric \citep{szalata2024benchmark,wu2024perturbench} is computed from any other metric, $m(\cdot,\cdot)$, by first measuring for the frequency with which the predicted perturbation, $\hat{\delta}_p$, is closer to the true perturbation, $\bar{\delta}_p$ than any other perturbation, $\bar{\delta}_q$, known as the rank. The rank is then averaged over all perturbations $p$ to produce the final retrieval statistic. Written definitions of these metrics can be found in Appendix \ref{ap:metric_definitions}


\begin{figure}[t!] 
    \centering
    \unitlength=1in 
    
    
    \begin{subfigure}[b]{0.31\textwidth}
        \centering
        \includegraphics[width=\linewidth]{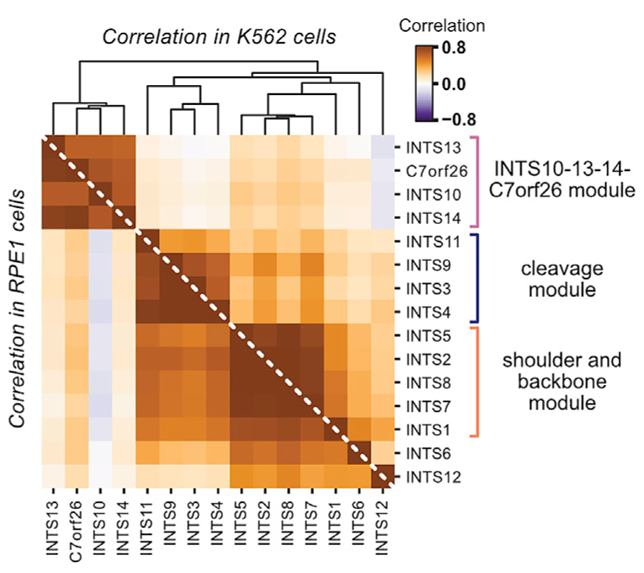}
        \label{fig:repogle_cross_correlation}
        \vspace{-2mm} 
    \end{subfigure}
    \hfill
    \begin{subfigure}[b]{0.32\textwidth}
        \centering
        \includegraphics[width=\linewidth]{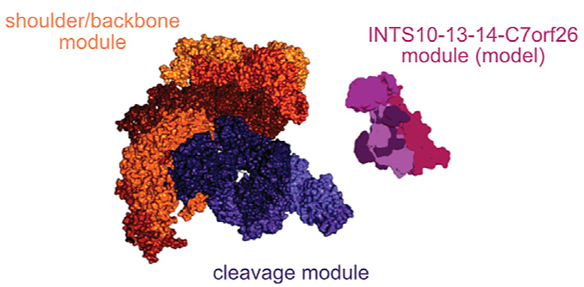}
        \label{fig:repogle_cluster}
        \vspace{-2mm}
    \end{subfigure}
    \hfill
    \begin{subfigure}[b]{0.32\textwidth}
        \centering
        \includegraphics[width=0.95\linewidth]{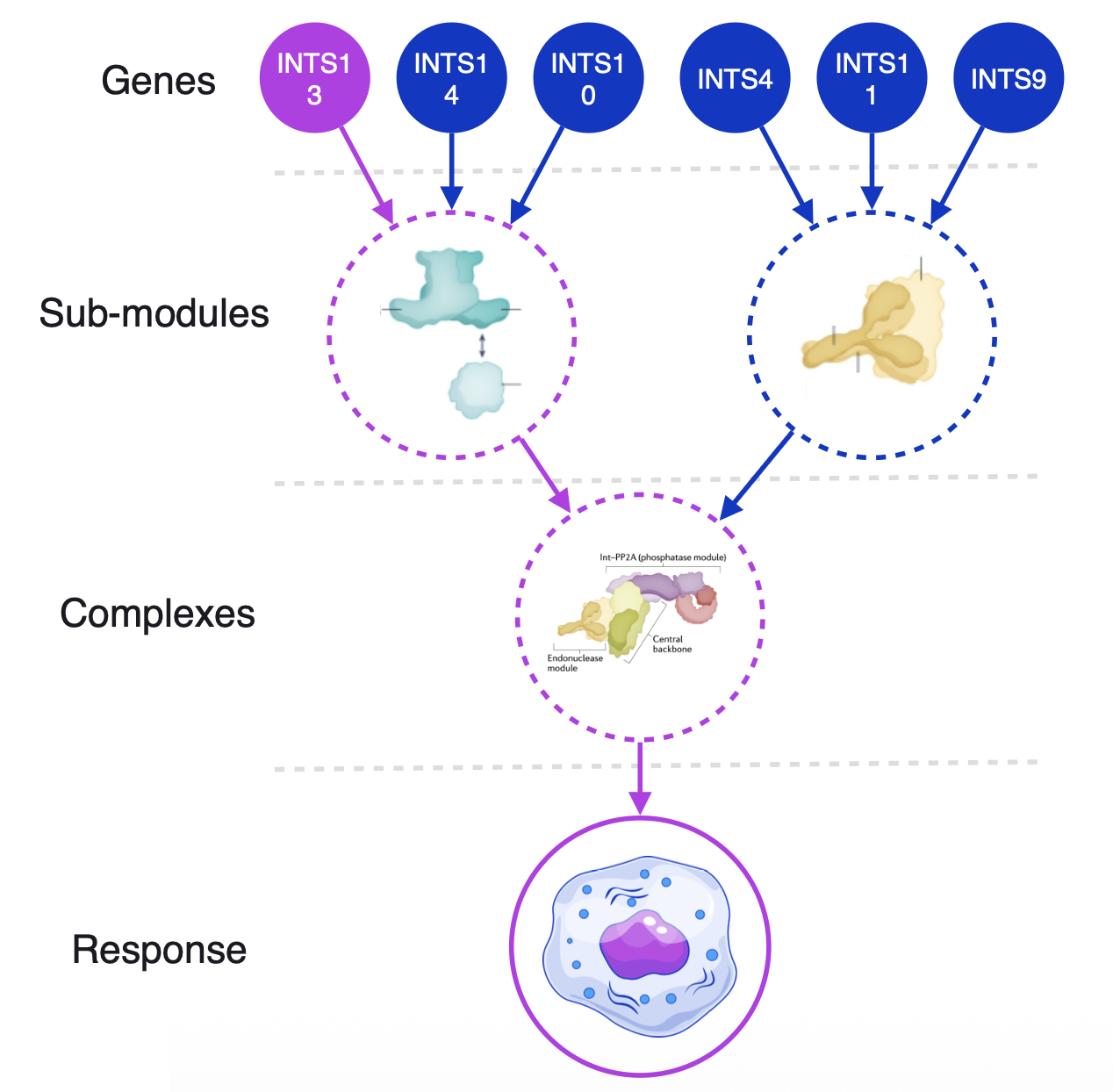}
        \label{fig:hierarchical_dag}
        \vspace{-2mm}
    \end{subfigure}

    \caption{\small \textbf{Hierarchical model of cell responses.} Our model abstracts the relationship between observed correlations (left) and underlying protein complexes (center) \citep{replogle2022mapping}. Genes in the same complex correlate highly; submodules are reflected in correlation blocks. This is represented as a causal DAG (right).}
    \label{fig:combined_figure}
\end{figure}
\section{A sparse hierachical model of cellular responses}
\label{sec:theory}

To motivate our choice of predictor, we first describe a simplistic hierarchical model of how gene knockouts relate to the cells response to a perturbation. Within cells, each gene provides the instructions to create a protein, different proteins come together to form complexes which fulfill different, larger roles in the cell, such as the ribosome or ATP synthase. Multiple complexes form pathways, which combine again to produce organelles like the nucleus and finally lead to the cell response. A gene knockout interrupts the production of the initial protein, which for our purpose we imagine to break all downstream levels of the hierarchy. 

This is clearly a simplification of biology, but it makes a useful testable prediction that perturbations of different genes will look more similar if they break lower levels of the same hierarchy (i.e both impacting the same complex) than if they affect unrelated pathways. As an analogy, if we were trying to understand the workings of a car by breaking different small pieces of it, the effects  of breaking parts of the suspension would be more similar to each other than to the effects of breaking a part that plays a role in the engine. This hierarchical model is consistent with some of the findings in \citet{replogle2022mapping} and presented in Figure \ref{fig:combined_figure}, where the cross cell line correlation between transcriptome signatures reveals submodules within the integrator complex.

We can formalize this assumption by assuming hierarchical Structural Causal Model (SCM) describes how gene knockouts affect the cell's response to a perturbation. We posit that the data generating process follows a Directed Acyclic Graph (DAG), $\mathcal{G}$, where nodes represent biological components at increasing levels of abstraction\footnote{Strictly speaking, some of the relationships in this DAG are better thought of as \emph{constituent} relationships rather than \emph{causal} relationships: submodels do not ``cause'' a complex any more than a table's legs ``cause'' a table. Instead a complex is constituted of sub-modules. }. Specifically, let the cell state be represented by a sequence of binary latent random vectors $\mathbf{Z} = (\mathbf{Z}_1, \dots, \mathbf{Z}_k)$, where $\mathbf{Z}_l \in \{0, 1\}^{n_l}$ represents the functional status ($1 = \text{active}, 0 = \text{broken}$) of biological components at level $l$ (e.g., from protein complexes at $l=1$ to high-level organelles at $l=k$). We then assume the following:
\begin{assumption}
    The transcriptomic response $\by$ is generated by $\bZ$ under the following structural constraints:
\begin{enumerate}[nosep, leftmargin=*, itemsep=0pt]
    \item \textbf{Failure Propagation:} $Z_{l+1, j} = \cap_{i \in \pa(Z_{l+1, j}) } Z_{l, i}$. The state of each $Z_{l+1, j}$ is causally determined by its parents, $\pa(Z_{l+1, j}) \subseteq \bZ_l$ and only remains functional if all of its parents are functional.
    \item \textbf{Perturbation Source:} A genetic perturbation $p$ affects only $\bZ_1$, specifically $\text{Ch}_1(p) \subset \bZ_1$. $\text{Ch}_k(p)$ denotes the causal descendants of $p$ in $\bZ_k$.
    \item \textbf{Response from Latents:} The observed expression $Y = F(\bZ)+\epsilon$ satisfies a \textit{layer-wise Lipschitz condition} with sensitivity constants $L_l$:
    \vspace{-5pt} 
    \begin{equation*}
        \| F(\bZ) - F(\bZ^{\prime}) \|_{2} \leq \sum_{i=1}^{K} L_l \| \bZ_i - \bZ_i^{\prime} \|_{1}
    \end{equation*}
\end{enumerate}
\end{assumption}

Based on these assumptions, we have that the difference between the expected response of two perturbations is bounded as follows:
\begin{equation*} \label{eq:bound}
    \textstyle \| \mathbb{E}[Y | \text{do}(p)] - \mathbb{E}[Y | \text{do}(p')] \|_2 
    \leq \sum_{i=1}^{K} L_l \| \text{Ch}_i(p) - \text{Ch}_i(p') \|_1
\end{equation*}
Therefore, if two distinct perturbations impact the same low level complex in $\bZ_1$, their interventional distributions would be equal,  $\mathbb{E}[Y | \text{do}(p)] = \mathbb{E}[Y | \text{do}(p')]$. Further, perturbations intersecting in the earlier level of the hierarchy would have more similar expression profiles than those intersecting later. Therefore, leveraging knowledge or similarities between genes, such as being in similar complexes, pathways, or having interacting proteins would themselves act as a strong prior for predictive performance. In the next section we show the simplest model to leverage this assumption, a K-nearest neighbour on a biological knowledge graph- can beat most models in the out-of-distribution perturbation prediction.

\section{The Surprising Effectiveness of Related Genes for Transcriptomic Prediction}\label{sec:gnn_baseline}

The simplest possible learning algorithm implied by the causal assumptions in the previous section is to predict a perturbation's effect as the nearest neighbour average of its immediate neighbours in the causal graph. While we do not know this causal graph, in this section we show that leveraging knowledge of related genes from known biological knowledge graphs leads to a very strong predictor of the effects of unseen pertubations. 

\begin{figure*}[t]
    \centering
    \includegraphics[width = \textwidth]{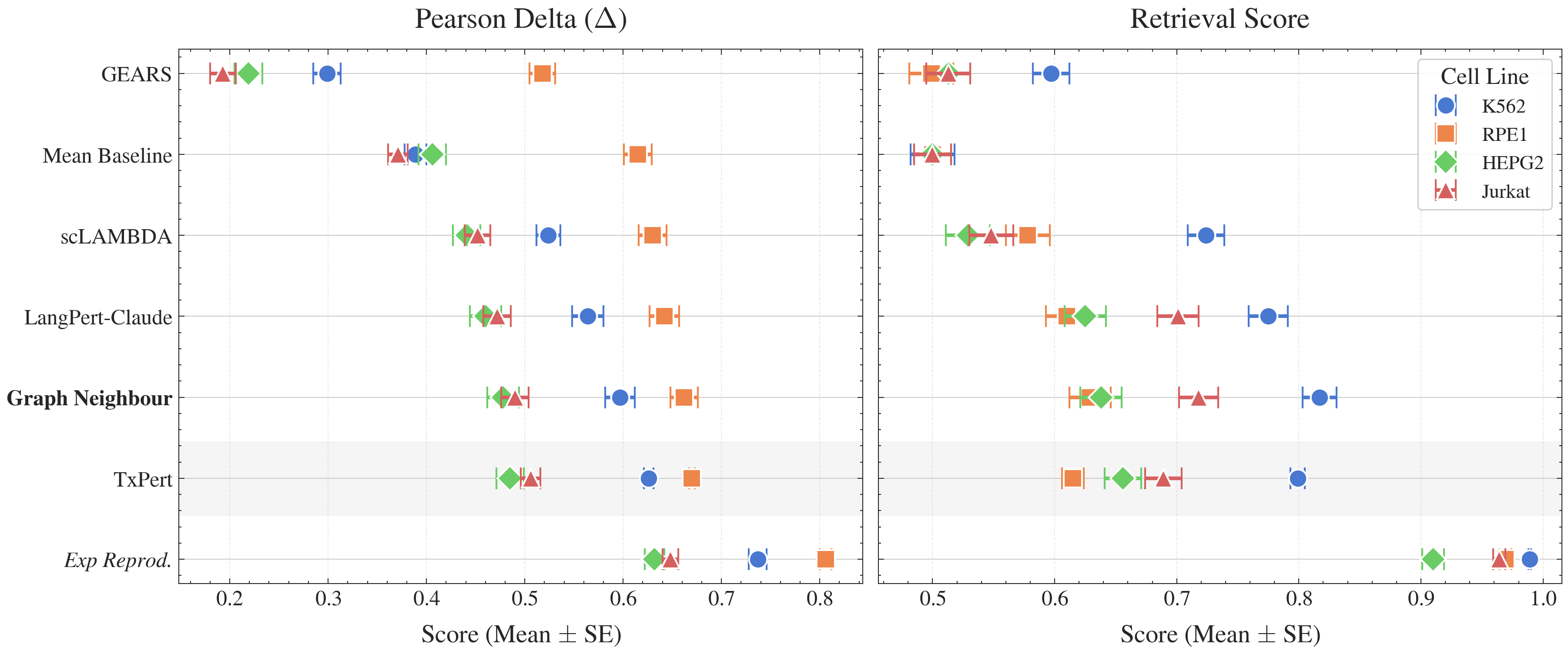}
    \caption{\small \textbf{Surprising Effectiveness of Related Genes} This figure compares the knowledge graph neighbour predictor on the cell lines from \citet{replogle2022mapping} against more complex perturbation prediction models in GEARS \citep{roohani2024predicting}, scLAMBDA \citep{wang2024modeling}, TxPert \citep{wenkel2025txpert} and an LLM based approach in Langpert \citep{martens2025langpert} using Claude Sonnet. We can see that the graph neighbour outperforms all but TxPert, a current state of the art model.}
    \label{fig:surprising_perf}
\end{figure*}
Empirically, we found that this model, which we term the \emph{graph neighbour predictor}, exhibited better predictive performance than most trained deep learning models, including GEARS \citep{roohani2024predicting} and scLAMBDA \citep{wang2024modeling}. 
This also explains the strong performance of LangPert \citep{martens2025langpert}, which uses a similar approach relying on an LLM to select the related genes. We found that related genes from a knowledge graph generally out perform those chosen by a base LLM and explains much of the LLM's predictive performance. In Section \ref{sec:LLM+RL} we will show how LLMs with access to knowledge graphs can be trained to find a nearest neighbour predictor that performs equivalently to recent state-of-the-art (SOTA) models \citep{wenkel2025txpert}


\subsection{Graph Neighbour Prediction}

Let $\cG =(\cV,\cE)$ be a biological knowledge graph where the vertices, $\cV$, are genes and the edges, $\cE$, are known relationships between genes, such as protein-protein interactions \citep{szklarczyk2023string}. For a gene, $g \in \cV$, let $N(g) \subseteq \cV $ be the set of neighbours in the knowledge graph and $N_{\mathrm{Tr}}(g) \coloneqq N(g) \cap \cP_{\mathrm{Tr}}$ be the intersection of this set with the training gene list. We then define the graph neighbour prediction for perturbing a gene, $g$, as:
\begin{equation}
    \hat{\mathbf{y}}_g = \bar{\mathbf{x}}_g + \frac{1}{|N_{\mathrm{Tr}}(g)|} \sum_{\tilde{g} \in N_{\mathrm{Tr}}(g)} \bar{\delta}_{\tilde{g}}.
\end{equation} 
For our knowledge graphs we either the concentration of the graphs from \citet{wenkel2025txpert} which consists of StringDB \citep{szklarczyk2023string}, GO \citep{gene2004gene}, and two graphs derived from perturbation screens. For each of these graphs, if the full neighbourhood size of a gene is above 20 we subset down based on weights of connection strength if they are available and randomly otherwise. Finally, we note that this is exactly the same form of prediction used in LangPert \citep{martens2025langpert}, however they query an LLM for the related gene set instead of a knowledge graph.


\subsection{The Strong Performance of the Graph Neighbour Predictor}

We evaluated the performance of the graph neighbour predictor on the 4 cell lines from \citet{replogle2022mapping}. To benchmark the performance of this model we included comparisons to LangPert using both Gemini 2.5 and Claude Sonnet 4 as the LLM, TxPert \citep{wenkel2025txpert} which is a state of the art model for these cell lines, the mean baseline, and a scaled version of the experimental reproducibility defined in the next paragraph. 

The experimental reproducibility for a test perturbation, $p$, is defined by splitting the dataset, $\cD_p$, of cells for that perturbation in half to get two new sets, $\cD_{p,1}$ and $\cD_{p,2}$, and using the mean expression in one set as the prediction of the mean of the other set. As this halves the size of the dataset used to estimate the mean, we include sample size correction, details in Appendix \ref{ap:scaled_exp_repro}.

Figure \ref{fig:surprising_perf}, shows that the graph neighbour model gave very competitive performance across all four cell lines, outperforming GEARS \citep{roohani2024predicting}, scLAMBDA \citep{wang2024modeling} and coming close to the performance of TxPert in Pearson delta. We include multiple versions of the graph neighbour with differing knowledge graphs, demonstrating the robustness of this result.

We also found that the effectiveness of related genes as perturbation predictors explains the Pearson Delta performance of LangPert \citep{martens2025langpert}. Specifically, there is significant overlap between genes selected by the knowledge graph and genes selected by the LLM in LangPert, between 30-50\% depending on model. Moreover, predicting using the intersection of knowledge graph neighbours and LangPert chosen genes lead to unchanged performance compared to LangPert, suggesting that the additional genes chosen by the LLM provided little to no benefit in predictive performance. Full analysis can be found in Appendix \ref{ap:langpert_vs_kg}.

\subsection{Empirical Analysis of the Graph Neighbour Predictor}\label{sec:empirical}

If the assumptions that we made in Section \ref{sec:theory} were correct, we knew the causal graph $\mathcal{G}$, and had access to a training set whose 1-hop neighbourhood covered the full graph, we could expect near perfect prediction from the graph neighbour approach. Whilst the graph neighbour predictor performs surprisingly well in practice---suggesting these assumptions hold for many genes---biological knowledge graphs are inherently incomplete approximations of $\mathcal{G}$. 
To understand how these imperfections affect performance, we empirically analyzed the graph neighbour predictor (Figure \ref{fig:three_points_main}; Appendix \ref{ap:graph_across_cell}), yielding three key findings:

\begin{itemize}[leftmargin=*, noitemsep, topsep=0pt]
    \item \textbf{Performance is near-optimal for a majority of genes.} For 60\% of genes, the predictor's Pearson delta has a very small gap to the experimental reproducibility ceiling. The overall performance gap is therefore driven by the tails of the distribution: the 60--90\% quantiles account for roughly two-thirds of the total gap, while the worst-performing 10\% of genes contribute the rest (Figure \ref{fig:exp_repro_vs_neighbour_main}).
    
    \item \textbf{Poorly predicted genes exhibit weaker, more ``control-like'' signals.} Perturbations that are poorly predicted generally have fewer training-set neighbours, though the large variance in neighbourhood size across groups (plotted in Figure \ref{fig:neighbourhood_size}) suggests this alone cannot explain the performance drop. We find that more important is that these genes have lower-magnitude perturbation deltas (Figure \ref{fig:perf_vs_mag}). This implies their perturbed state is closer on average to unperturbed cells, and therefore the predictive performance strongly correlates with how well the relevant controls predict the outcome \textit{(see Appendix \ref{ap:performance_corr} for this correlation, and its reversal for MSE and MAE metrics)}.
    
    \item \textbf{Knowledge graph neighbourhoods are strong but fundamentally suboptimal.} By comparing the distribution of Pearson deltas between a given gene and the full training set versus just its KG neighbours, we see that while KG neighbours are highly predictive on average, the immediate neighbourhood still misses key high-performing genes and includes genes with low predictive power (Figure \ref{fig:violin_hepg2}).
\end{itemize}

Because KGs are built on known biological relationships rather than raw predictive power, these neighbour sets are not strictly optimal. This leaves significant room for improvement, motivating our search over sets using an LLM trained via RL in the next section.

\section{Reasoning LLMs to Find Better Neighbourhoods}\label{sec:LLM+RL}
\begin{figure}[t] 
    \centering
    \small 
    
    \begin{subfigure}[b]{0.32\linewidth}
        \centering
        \includegraphics[width=\linewidth]{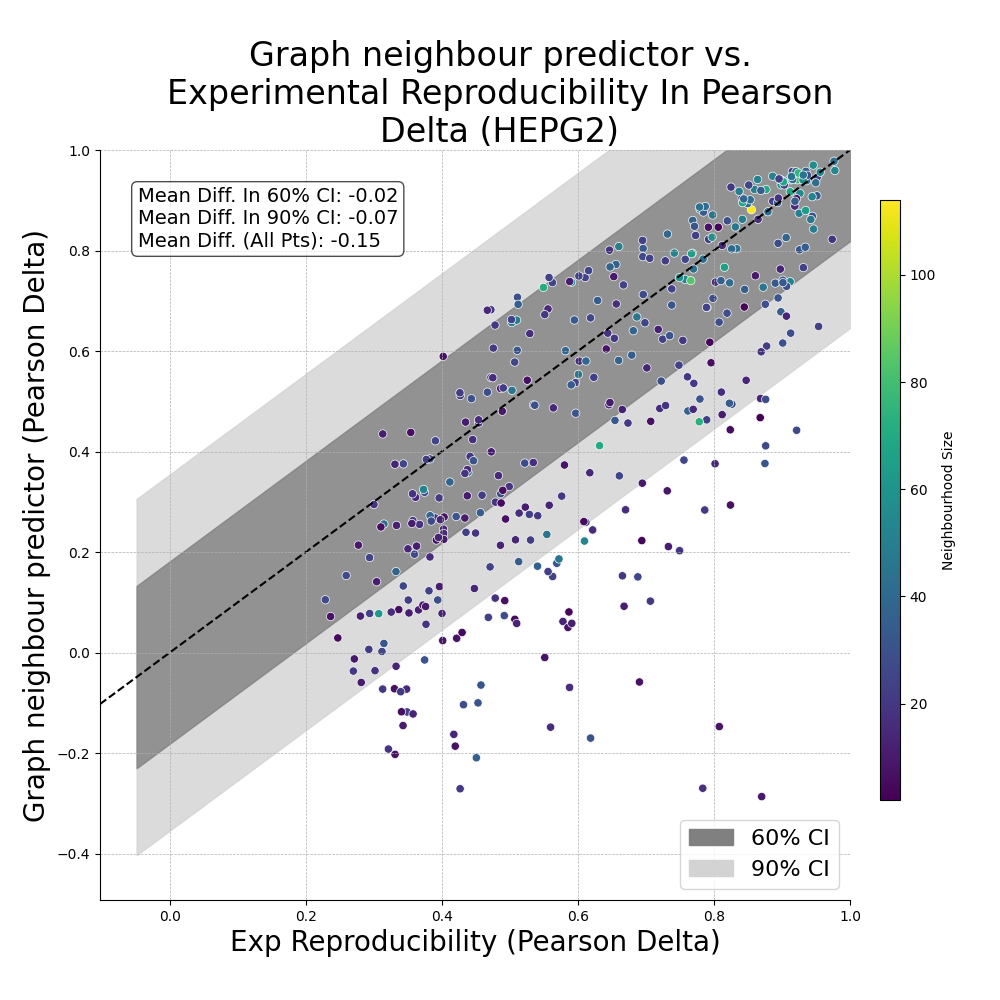}
        \vspace{-6pt}
        \caption{Predictor vs. Repro.}
        \label{fig:exp_repro_vs_neighbour_main}
    \end{subfigure}
    \hfill
    \begin{subfigure}[b]{0.32\linewidth}
        \centering
        \includegraphics[width=\linewidth]{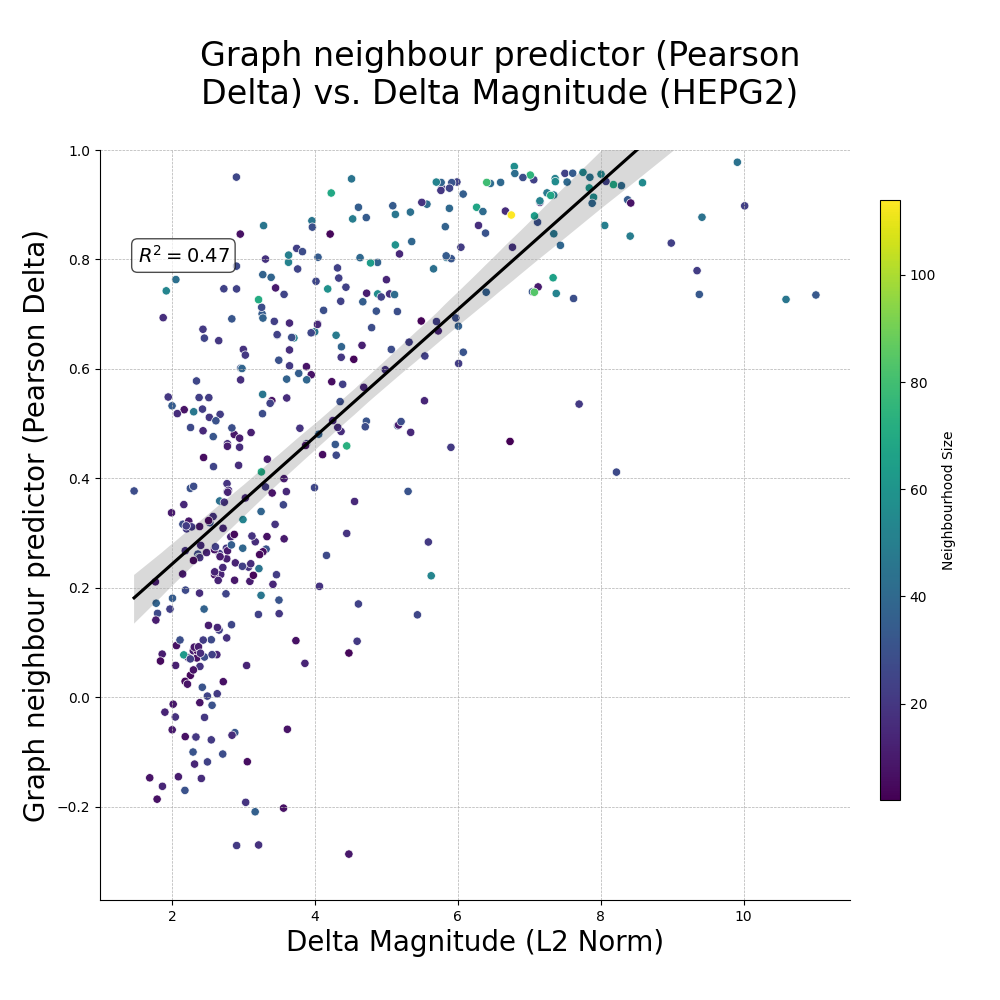}
        \vspace{-6pt}
        \caption{Perf. vs. Magnitude $\delta$.}
        \label{fig:perf_vs_mag}
    \end{subfigure}
    \hfill
    \begin{subfigure}[b]{0.32\linewidth}
        \centering
        \includegraphics[width=\linewidth]{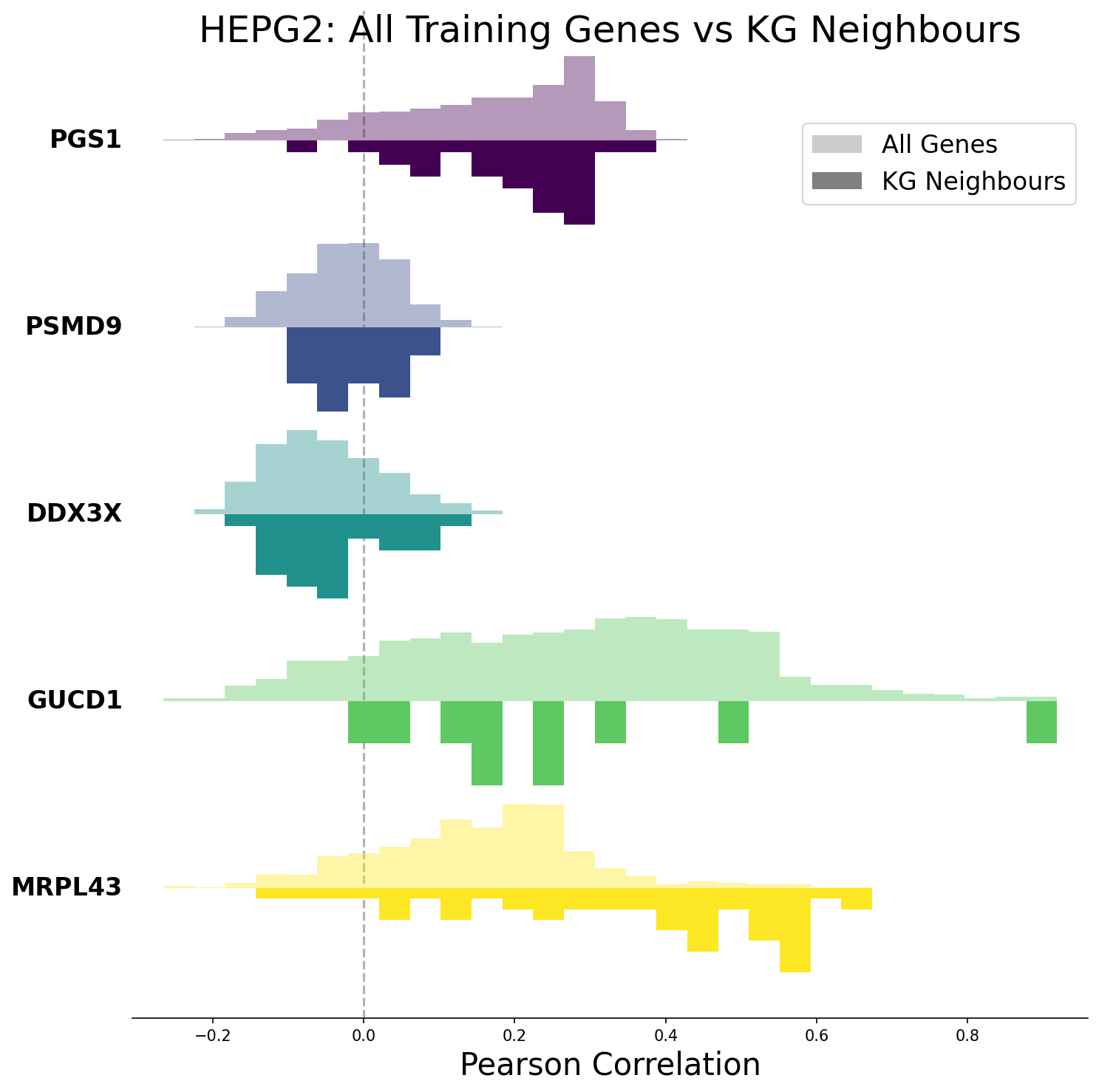}
        \vspace{-6pt}
        \caption{Correlation Distribution.}
        \label{fig:violin_hepg2}
    \end{subfigure}

    \caption{\textbf{Graph neighbour predictor performance.} \textbf{(a)} Our predictor is competitive with experimental reproducibility (Pearson $\Delta$); shaded areas denote 60\% and 90\% performance gap quantiles, which account for majority of the gap. \textbf{(b)} Model performance correlates strongly with perturbation magnitude, meaning more discernible perturbations are easier to predict. \textbf{(c)} Knowledge graph neighbourhoods capture high-correlation genes but may omit specific related perturbations. This plot demonstrates this by visualizing the predictive performance of the neighbourhood vs the whole training set in Pearson delta.}
    \label{fig:three_points_main}
\end{figure}
In Section 2, we demonstrated that the set of gene neighbours in a biological knowledge graph, $N_{\mathrm{KG}}(p)$, provides a strong baseline for transcriptomic perturbation effect prediction. However, we know from the incompleteness of knowledge graphs and our empirical analysis that these gene sets are not optimal. To refine these neighbourhoods, we would ideally like to optimize over the space of all possible graphs that connect these genes. But search over $O(2^{|\text{genes}|^2})$ graphs is intractible.
Instead we search over this space using LLMs trained via reinforcement learning to maximize predictive performance. In doing this we implicitly constrain our search to potential genes that are likely under the LLM's prior beliefs over relationships between genes. Through training the LLM is able to refine a candidate biological neighbourhood, effectively learning a generalizable policy for selecting the sets of predictive genes.

\subsection{Prediction Method}

Given the previous section, we formulate the prediction task as one of selecting a set of related genes, $S$, where the predicted delta, denoted $\delta(S)$, is calculated as the mean of the training deltas, $\bar{\delta}_q$, for genes, $q\in S$. The final predicted expression profile, $\hat{\mathbf{y}}_p$, is then obtained by adding the mean of paired control, $\bar{\mathbf{x}}_p$. Bringing this together: 
\begin{align*}
    \delta(S_{\mathrm{LLM}}) = \frac{1}{|S_{\mathrm{LLM}}|} \sum_{q \in S_{\mathrm{LLM}}} \bar{\delta}_q, \quad \quad \hat{\mathbf{y}}_p = \bar{\mathbf{x}}_p + \delta(S).
\end{align*}

We then aim to get an LLM to select the predictive set, similarly to LangPert \citep{martens2025langpert}. However,
given the strong performance of the graph neighbour baseline, we start by giving the LLM  the immediate graph neighbours, $N_{\mathrm{KG}}(p)$, and ask it to propose a set of additions, $A_{\mathrm{LLM}}$, and removals, $R_{\mathrm{LLM}}$. The final set, $S_{\mathrm{LLM}}$, is then formed as: $S_{\mathrm{LLM}} = \left( N_{\mathrm{KG}}(p) \setminus R_{\mathrm{LLM}} \right) \cup A_{\mathrm{LLM}}$.

We also provide the LLM with additional context for the perturbation and cell line it is trying to predict. Full prompts are given in Appendix \ref{ap:RL_LLLM_prompt}. We note that the LLM often proposes additions or removals that are not possible, such as genes not present in the training set or removals of genes not in the neighbourhood.. When this happens we ignore these and take all available additions/removals. If the final set is empty we predict the mean of all training perturbations. 

\subsection{Reinforcement Learning Setup}\label{subsec:RL}

To train the LLM to discover generalizable reasoning patterns for set selection, we optimize the policy $\pi_\theta$ using GRPO \citep{shao2024deepseekmath}. For a prompt, $\mathbf{s}$, with associated perturbation $p$, and a completion $\mathbf{a}$ containing the actions $A_{\mathrm{LLM}}(\mathbf{a})$ and $R_{\mathrm{LLM}}(\mathbf{a})$, this reward is the mean of three components: $R(\mathbf{s},\mathbf{a}) = \frac{1}{3} \left( R_{\mathrm{Fmt}}(\mathbf{s},\mathbf{a}) + R_{\mathrm{Valid}}(\mathbf{s},\mathbf{a}) + R_{\Delta}(\mathbf{s},\mathbf{a}) \right)$. The first, $R_{\mathrm{Fmt}}$, is a formatting reward to ensure the sets can be extracted from the output. The other two are as:

\textbf{Validity Reward ($R_{\mathrm{Valid}}$):} encourages the LLM to suggest adding genes in the training set $\mathcal{P}_{\mathrm{Tr}}$ and removing genes in the original neighbourhood $N_{\mathrm{KG}}(p)$:
\begin{equation*}
    \textstyle R_{\mathrm{Val}}(\bs,\ba) = \frac{|A_{\mathrm{LLM}} \cap (\mathcal{P}_{\mathrm{Tr}} \setminus \{p\})| + |R_{\mathrm{LLM}} \cap N_{\mathrm{KG}}(p)|}{|A_{\mathrm{LLM}}| + |R_{\mathrm{LLM}}|}
\end{equation*}
If no changes are proposed ($|A_{\mathrm{LLM}}| + |R_{\mathrm{LLM}}| = 0$), we set $R_{\mathrm{Valid}} = 0$ in order to incentivise the model to alter the neighbourhood.

\textbf{Pearson Delta Reward ($R_{\Delta}$):} Finally, this reward incentivises the model to find better predictive sets in terms of Pearson delta. Specifically, let $\Delta_{\mathrm{diff}}=\Delta(\hat{\delta}_{\mathrm{LLM}}, \bar{\delta}_p) - \Delta(\hat{\delta}_{\mathrm{KG}}, \bar{\delta}_p)$ be the performance gap between the LLM set and the knowledge graph. We then set $R_{\Delta}(\mathbf{s},\mathbf{a}) = \lambda\Delta_{\mathrm{diff}} + \mathbb{I}\left[\Delta_{\mathrm{diff}} > \epsilon \right]$ where $\epsilon, \lambda$ are hyper-parameters and the second term provides a consistent reward signal as $\Delta_{\mathrm{diff}}$ is very noisy and can vary in scale from gene to gene.


\subsection{Training}

Throughout, we maintain a train/test split so that test perturbations and their knowledge graphs remain unseen to the LLM. The question set is formed by using all training perturbations, randomly shuffled and not grouped by cell line. We use Qwen3-4B-Instruct-2507 as the base model and Prime-RL \citep{primeintellect2025prime-rl} for training, details in Appendix \ref{ap:rl_train}.
\begin{table*}[t!]
\centering
\scriptsize 
\setlength{\tabcolsep}{2.5pt} 

\newcommand{\res}[2]{$#1_{\pm #2}$}
\setlength{\floatsep}{5pt}           
\setlength{\textfloatsep}{5pt}
\resizebox{\textwidth}{!}{
\newcommand{\bres}[2]{\textbf{#1} \tiny(\textbf{#2})}

\begin{tabular}{l cc cc cc cc}
\toprule
& \multicolumn{2}{c}{\textbf{K562}} & \multicolumn{2}{c}{\textbf{RPE1}} & \multicolumn{2}{c}{\textbf{HEPG2}} & \multicolumn{2}{c}{\textbf{Jurkat}} \\
\cmidrule(lr){2-3} \cmidrule(lr){4-5} \cmidrule(lr){6-7} \cmidrule(lr){8-9}
\textbf{Model} & Pear. $\Delta (\uparrow)$ & Retr. $(\uparrow)$ & Pear. $\Delta (\uparrow)$ & Retr. $(\uparrow)$ & Pear. $\Delta (\uparrow)$ & Retr. $(\uparrow)$ & Pear. $\Delta (\uparrow)$ & Retr. $(\uparrow)$ \\ 
\midrule
Exp Reprod. & \res{0.737}{.009} & \res{0.989}{.001} & \res{0.806}{.006} & \res{0.969}{.005} & \res{0.632}{.010} & \res{0.910}{.009} & \res{0.648}{.008} & \res{0.964}{.005} \\ 
\midrule
LLM, RL, + KG (\textbf{Ours}) & \bres{0.625}{.014} & \bres{0.847}{.013} & \bres{0.672}{.014} & \res{0.649}{.017} & \bres{0.486}{.016} & \res{0.656}{.017} & \bres{0.510}{.014} & \res{0.737}{.016} \\
LLM+KG & \res{0.599}{.015} & \res{0.823}{.014} & \res{0.659}{.014} & \res{0.641}{.017} & \res{0.473}{.016} & \res{0.650}{.016} & \res{0.497}{.014} & \res{0.738}{.016} \\
KG & \res{0.597}{.015} & \res{0.817}{.014} & \res{0.662}{.014} & \res{0.629}{.017} & \res{0.478}{.016} & \res{0.638}{.017} & \res{0.490}{.014} & \res{0.718}{.016} \\ 
TxPert & \bres{0.626}{.005} & \res{0.799}{.006} & \bres{0.670}{.003} & \res{0.615}{.009} & \bres{0.485}{.014} & \res{0.656}{.015} & \bres{0.506}{.010} & \res{0.689}{.015} \\ 
\bottomrule
\end{tabular}
}
\caption{\textbf{Performance of LLM trained via RL with Knowledge Graph input} lines. Scores are Mean$_{\pm \text{se}}$. Best performing models (excluding Exp Reproducibility) are highlighted in \textbf{bold} with statistical significance established via  pairwise Wilcoxon signed-rank tests. LLM +KG uses a comparable frontier LLM (Gemini 2.5 pro) which outperforms the untrained base model.}
\label{tab:model_performance_flat}
\end{table*}
\section{Results}

We now provide our results for the LLM trained using the procedure in Section \ref{sec:LLM+RL} on the \citet{replogle2022mapping} cell lines, showing our model can find a ``nearest neighbour'' predictor which improves over the graph neighbour method introduced in Section \ref{sec:gnn_baseline} and performs similarly to a current state of the art model \citep{wenkel2025txpert}. We then perform additional experiments, aiming to understand its generalisation across cell lines and assessing the post-trained LLM on downstream tasks. Notably we find that the post-trained LLM exhibits better differential expression prediction on PerturbQA \citep{wu2024perturbench} than the base model. 

\subsection{Unseen Perturbation Prediction}

Firstly, we assess on unseen perturbation prediction, using the same train/test splits as in Section \ref{sec:gnn_baseline} from \citet{wenkel2025txpert}. Our results can be see in Table \ref{tab:model_performance_flat}, where we compare the RL trained LLM with all the models discussed in Section \ref{sec:gnn_baseline}, as well as giving an untrained frontier LLM (Gemini 2.5 Pro) the same prompts to edit the neighbourhoods. As the difference between models is small and the per-cell line test datasets are also small, we use pairwise Wilcoxon signed-rank tests to establish the difference between models..

Our results show that the LLM trained via RL is able to outperform the graph neighbour predictor across all four cell lines, with equivalent performance to TxPert \citep{wenkel2025txpert}. Notably, we also find that the frontier LLM does improve over the graph neighbour predictor, suggesting that the improvements of our model are not solely driven by ``obvious'' changes to the neighbourhoods.

\begin{figure*}[t]
    \centering
    \includegraphics[width = \textwidth]{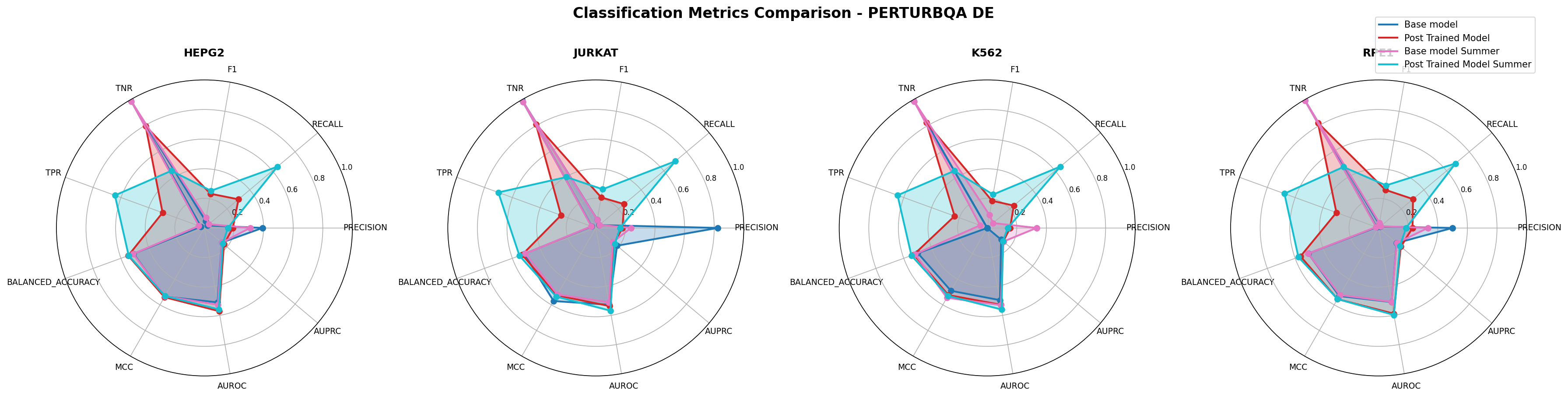}
    \caption{\small \textbf{Differential expression prediction performance.} The post-trained model consistently outperforms the base model in the tasks from \citet{wu2025contextualizing} with and without SUMMER-based prompting.}
    \label{fig:exp_repro}
\end{figure*}

\subsection{Transfer In Reasoning Across Cell Lines}

In order to assess how the reasoning learnt by the LLM generalises across cell lines, we repeat the same task training on only 3 of the 4 cell lines. We choose to leave out K562 as this is cell line on which the models performance improved most, making it easier to pinpoint drops in performance.  We stop the model after 750 steps, which represents half of the total training steps. Results are shown in table \ref{tab:cell_line_subset}.
\begin{wraptable}{r}{0.38\textwidth}
    \newcommand{\res}[2]{$#1_{\pm #2}$}
    \centering
    \footnotesize 
    \setlength{\tabcolsep}{0pt} 
    \begin{tabular*}{\linewidth}{@{\extracolsep{\fill}} lcc @{}} 
    \toprule
    \textbf{Model} & Pear. $\Delta \uparrow$ & Retr. $\uparrow$ \\ 
    \midrule
    \rowcolor[gray]{.95} \multicolumn{3}{l}{\textit{K562}} \\
    4 Cell Lines & \res{\mathbf{0.617}}{\mathbf{0.014}} & \res{0.840}{0.013} \\
    3 Cell Lines & \res{0.608}{0.015} & \res{0.834}{0.013} \\
    Base         & \res{0.593}{.015} & \res{0.813}{.014} \\
    \midrule
    \rowcolor[gray]{.95} \multicolumn{3}{l}{\textit{Jurkat}} \\
    4 Cell Lines & \res{\mathbf{0.504}}{\mathbf{0.014}} & \res{0.734}{0.016} \\
    3 Cell Lines & \res{0.496}{0.014} & \res{0.729}{0.016} \\
    Base         & \res{0.486}{.014} & \res{0.714}{.016} \\
    \bottomrule
    \end{tabular*}
    \caption{Comparing the performance of the RL and LLM approach when trained on all cell lines vs not trained on the K562 cell line after 750 steps alongside the base model. We can see that even when not trained on K562 the RL training improves the models performance over the base model.}
    \label{tab:cell_line_subset}
    \vspace{-15pt} 
\end{wraptable}
We find two key points: i) the LLM is able to exhibit some generalisation across cell line, where training across the other 3 cell lines still improves performance compared to the base model and knowledge graph on K562. ii) Removing training on K562 decreases performance on all 4 cell lines, even though the model has trained for longer on them.

\subsection{Performance on Downstream tasks}
Having demonstrated that RL training an LLM can allow us to find strong out of distribution perturbation predictors, we now turn to ask how the RL training changes the LLM. Specifically, we assess its performance on the downstream task of PerturbQA \citep{wucontextualizing}. PerturbQA consists of a variety of tasks, of which we focus on two in particular, differential expression prediction and direction of change. In differential expression, the model is given a pair of genes and asked if a perturbation on the first gene will lead to a statistically significant change in the expression levels of the second gene.  In the direction of change the model is again given a pair of genes, told there is statistically significant change, and then asked to predict which direction this occurs in.\citet{wucontextualizing} introduce this task and propose SUMMER, a system that combines RAG with structured prompting  to enhance performance.

We assess the performance of our post-trained model both with and without SUMMER prompting. In both cases we find that the RL trained model outperforms the base model on differential expression prediction, demonstrating signs that the post training transfers reasoning to downstream tasks. As PerturbQA is constructed on the datasets from \citet{replogle2022mapping}, we maintain the same train test split, ensuring we are not assessing the model on genes for which it has seen the knowledge graph. 

\section{Conclusion and Limitations} In this work, we demonstrated that simple graph nearest neighbour predictors are surprisingly strong baselines for unseen perturbation prediction. Further we showed that LLMs can be used as RL tuneable priors over related genes  to make this simple predictor state-of-the-art level. However, there are some limitations. First, applying RL to language models is much more computationally expensive than many existing methods in this space. We therefore view this not as the practical method of choice for unseen perturbation prediction but as a proof of concept that LLMs can provide meaningful predictive signal beyond knowledge graphs. Secondly, performance is dependent on the number of related genes and having a training set for that cell line, preventing "zero-shot" predictions in novel cell lines. We note this limitation is shared by many of the evaluated baselines.

\bibliography{references}

\begin{thebibliography}{38}
\providecommand{\natexlab}[1]{#1}
\providecommand{\url}[1]{\texttt{#1}}
\expandafter\ifx\csname urlstyle\endcsname\relax
  \providecommand{\doi}[1]{doi: #1}\else
  \providecommand{\doi}{doi: \begingroup \urlstyle{rm}\Url}\fi

\bibitem[Ahlmann-Eltze et~al.(2025)Ahlmann-Eltze, Huber, and Anders]{AhlmannEltze2025}
C.~Ahlmann-Eltze, W.~Huber, and S.~Anders.
\newblock Deep-learning-based gene perturbation effect prediction does not yet outperform simple linear baselines.
\newblock \emph{Nature Methods}, 22\penalty0 (8):\penalty0 1657--1661, 2025.
\newblock \doi{10.1038/s41592-025-02772-6}.
\newblock URL \url{https://doi.org/10.1038/s41592-025-02772-6}.

\bibitem[Alberts et~al.(2002)Alberts, Johnson, Lewis, Raff, Roberts, and Walter]{Alberts2002}
B.~Alberts, A.~Johnson, J.~Lewis, M.~Raff, K.~Roberts, and P.~Walter.
\newblock \emph{Molecular Biology of the Cell}.
\newblock Garland Science, New York, NY, 4th edition, 2002.
\newblock ISBN 9780815332181.

\bibitem[Bendidi et~al.()Bendidi, Whitfield, Kenyon-Dean, Yedder, El~Mesbahi, Noutahi, and Denton]{bendidi2024benchmarking}
I.~Bendidi, S.~T. Whitfield, K.~Kenyon-Dean, H.~B. Yedder, Y.~El~Mesbahi, E.~Noutahi, and A.~K. Denton.
\newblock Benchmarking transcriptomics foundation models for perturbation analysis: one pca still rules them all.
\newblock In \emph{NeurIPS 2024 Workshop on AI for New Drug Modalities}.

\bibitem[Brixi et~al.(2025)Brixi, Durrant, Ku, Poli, Brockman, Chang, Gonzalez, King, Li, Merchant, et~al.]{brixi2025genome}
G.~Brixi, M.~G. Durrant, J.~Ku, M.~Poli, G.~Brockman, D.~Chang, G.~A. Gonzalez, S.~H. King, D.~B. Li, A.~T. Merchant, et~al.
\newblock Genome modeling and design across all domains of life with evo 2.
\newblock \emph{BioRxiv}, pages 2025--02, 2025.

\bibitem[Brown(2025)]{brown_verifiers_2025}
W.~Brown.
\newblock {Verifiers}: Environments for llm reinforcement learning.
\newblock \url{https://github.com/willccbb/verifiers}, 2025.

\bibitem[Bunne et~al.(2023)Bunne, Stark, Gut, del Castillo, Levesque, Lehmann, Pelkmans, Krause, and R{\"a}tsch]{Bunne2023aa}
C.~Bunne, S.~G. Stark, G.~Gut, J.~S. del Castillo, M.~Levesque, K.-V. Lehmann, L.~Pelkmans, A.~Krause, and G.~R{\"a}tsch.
\newblock Learning single-cell perturbation responses using neural optimal transport.
\newblock \emph{Nature Methods}, 20\penalty0 (11):\penalty0 1759--1768, 2023.
\newblock \doi{10.1038/s41592-023-01969-x}.
\newblock URL \url{https://doi.org/10.1038/s41592-023-01969-x}.

\bibitem[Bunne et~al.(2024)Bunne, Roohani, Rosen, Gupta, Zhang, Roed, Alexandrov, AlQuraishi, Brennan, Burkhardt, Califano, Cool, Dernburg, Ewing, Fox, Haury, Herr, Horvitz, Hsu, Jain, Johnson, Kalil, Kelley, Kelley, Kreshuk, Mitchison, Otte, Shendure, Sofroniew, Theis, Theodoris, Upadhyayula, Valer, Wang, Xing, Yeung-Levy, Zitnik, Karaletsos, Regev, Lundberg, Leskovec, and Quake]{bunne2024}
C.~Bunne, Y.~Roohani, Y.~Rosen, A.~Gupta, X.~Zhang, M.~Roed, T.~Alexandrov, M.~AlQuraishi, P.~Brennan, D.~Burkhardt, A.~Califano, J.~Cool, A.~Dernburg, K.~Ewing, E.~Fox, M.~Haury, A.~Herr, E.~Horvitz, P.~Hsu, V.~Jain, G.~Johnson, T.~Kalil, D.~Kelley, S.~Kelley, A.~Kreshuk, T.~Mitchison, S.~Otte, J.~Shendure, N.~Sofroniew, F.~Theis, C.~Theodoris, V.~S. Upadhyayula, M.~Valer, B.~Wang, E.~Xing, S.~Yeung-Levy, M.~Zitnik, T.~Karaletsos, A.~Regev, E.~Lundberg, J.~Leskovec, and S.~Quake.
\newblock How to build the virtual cell with artificial intelligence: Priorities and opportunities.
\newblock \emph{Cell}, 187\penalty0 (25):\penalty0 7045--7063, 2024.
\newblock \doi{10.1016/j.cell.2024.11.015}.

\bibitem[Chen and Zou(2024)]{genept}
Y.~Chen and J.~Zou.
\newblock Genept: A simple but effective foundation model for genes and cells built from chatgpt.
\newblock \emph{bioRxiv}, Mar 2024.
\newblock ISSN 2692-8205 (Electronic); 2692-8205 (Linking).
\newblock \doi{10.1101/2023.10.16.562533}.

\bibitem[Consortium(2004)]{gene2004gene}
G.~O. Consortium.
\newblock The gene ontology (go) database and informatics resource.
\newblock \emph{Nucleic acids research}, 32\penalty0 (suppl\_1):\penalty0 D258--D261, 2004.

\bibitem[Cui et~al.(2024)Cui, Wang, Maan, Pang, Luo, Duan, and Wang]{Cui2024}
H.~Cui, C.~Wang, H.~Maan, K.~Pang, F.~Luo, N.~Duan, and B.~Wang.
\newblock scgpt: toward building a foundation model for single-cell multi-omics using generative ai.
\newblock \emph{Nature Methods}, 21\penalty0 (8):\penalty0 1470--1480, 2024.
\newblock \doi{10.1038/s41592-024-02201-0}.
\newblock URL \url{https://doi.org/10.1038/s41592-024-02201-0}.

\bibitem[Fallahpour et~al.(2025)Fallahpour, Magnuson, Gupta, Ma, Naimer, Shah, Duan, Ibrahim, Goodarzi, Maddison, and WANG]{fallahpour2025bioreason}
A.~Fallahpour, A.~Magnuson, P.~Gupta, S.~Ma, J.~Naimer, A.~Shah, H.~Duan, O.~Ibrahim, H.~Goodarzi, C.~J. Maddison, and B.~WANG.
\newblock Bioreason: Incentivizing multimodal biological reasoning within a {DNA}-{LLM} model.
\newblock In \emph{The Thirty-ninth Annual Conference on Neural Information Processing Systems}, 2025.
\newblock URL \url{https://openreview.net/forum?id=mDjEKAwJOF}.

\bibitem[Gaudelet et~al.(2024)Gaudelet, Vecchio, Carrami, Cudini, Kapourani, Uhler, and Edwards]{gaudelet2024seasoncombinatorialinterventionpredictions}
T.~Gaudelet, A.~D. Vecchio, E.~M. Carrami, J.~Cudini, C.-A. Kapourani, C.~Uhler, and L.~Edwards.
\newblock Season combinatorial intervention predictions with salt \& peper, 2024.
\newblock URL \url{https://arxiv.org/abs/2404.16907}.

\bibitem[Intellect(2025)]{primeintellect2025prime-rl}
P.~Intellect.
\newblock Prime-rl, 2025.
\newblock URL \url{https://github.com/PrimeIntellect-ai/prime-rl}.

\bibitem[Istrate et~al.(2025)Istrate, Milletari, Castrotorres, Tomczak, Torkar, Li, and Karaletsos]{istrate2025rbio1}
A.-M. Istrate, F.~Milletari, F.~Castrotorres, J.~M. Tomczak, M.~Torkar, D.~Li, and T.~Karaletsos.
\newblock rbio1-training scientific reasoning llms with biological world models as soft verifiers.
\newblock \emph{bioRxiv}, pages 2025--08, 2025.

\bibitem[Korbak et~al.(2022)Korbak, Perez, and Buckley]{korbak2022rl}
T.~Korbak, E.~Perez, and C.~Buckley.
\newblock Rl with kl penalties is better viewed as bayesian inference.
\newblock In \emph{Findings of the Association for Computational Linguistics: EMNLP 2022}, pages 1083--1091, 2022.

\bibitem[Levine et~al.(2024)Levine, Rizvi, L{\'e}vy, Pallikkavaliyaveetil, Zhang, Chen, Ghadermarzi, Wu, Zheng, Vrkic, et~al.]{levine2024cell2sentence}
D.~Levine, S.~A. Rizvi, S.~L{\'e}vy, N.~Pallikkavaliyaveetil, D.~Zhang, X.~Chen, S.~Ghadermarzi, R.~Wu, Z.~Zheng, I.~Vrkic, et~al.
\newblock Cell2sentence: teaching large language models the language of biology.
\newblock \emph{BioRxiv}, pages 2023--09, 2024.

\bibitem[Liu et~al.(2025)Liu, Zhang, Du, Zhao, and Wang]{liu2025effects}
Q.~Liu, Q.~Zhang, J.~Du, S.~Zhao, and J.~Wang.
\newblock Effects of distance metrics and scaling on the perturbation discrimination score.
\newblock \emph{arXiv preprint arXiv:2511.16954}, 2025.

\bibitem[Lotfollahi et~al.(2019)Lotfollahi, Wolf, and Theis]{lotfollahi2019scgen}
M.~Lotfollahi, F.~A. Wolf, and F.~J. Theis.
\newblock scgen predicts single-cell perturbation responses.
\newblock \emph{Nature Methods}, 16\penalty0 (8):\penalty0 715--721, 2019.
\newblock \doi{10.1038/s41592-019-0494-8}.

\bibitem[Lotfollahi et~al.(2023)Lotfollahi, Naghipourfar, Luecken, Khajavi, B{"u}ttner, Avsec, Gayoso, Yosef, Theis, and Lopez]{lotfollahi2023cpa}
M.~Lotfollahi, M.~Naghipourfar, M.~D. Luecken, M.~Khajavi, M.~B{"u}ttner, {\v{Z}}.~Avsec, A.~Gayoso, N.~Yosef, F.~J. Theis, and R.~Lopez.
\newblock Predicting cellular responses to complex perturbations in high-throughput screens.
\newblock \emph{Molecular Systems Biology}, 19\penalty0 (5):\penalty0 e11517, 2023.
\newblock \doi{10.15252/msb.202211517}.

\bibitem[M{\"a}rtens et~al.()M{\"a}rtens, Martell, Prada-Medina, and Donovan-Maiye]{martens2025langpert}
K.~M{\"a}rtens, M.~B. Martell, C.~A. Prada-Medina, and R.~Donovan-Maiye.
\newblock Langpert: Llm-driven contextual synthesis for unseen perturbation prediction.
\newblock In \emph{ICLR 2025 Workshop on Machine Learning for Genomics Explorations}.

\bibitem[Murphy(2024)]{murphy2024reinforcement}
K.~Murphy.
\newblock Reinforcement learning: an overview.
\newblock \emph{arXiv preprint arXiv:2412.05265}, 2024.

\bibitem[Noutahi et~al.(2025)Noutahi, Hartford, Tossou, Whitfield, Denton, Wognum, Ulicna, Craig, Hsu, Cuccarese, et~al.]{noutahi2025virtual}
E.~Noutahi, J.~Hartford, P.~Tossou, S.~Whitfield, A.~K. Denton, C.~Wognum, K.~Ulicna, M.~Craig, J.~Hsu, M.~Cuccarese, et~al.
\newblock Virtual cells: Predict, explain, discover.
\newblock \emph{arXiv preprint arXiv:2505.14613}, 2025.

\bibitem[Replogle et~al.(2022)Replogle, Saunders, Pogson, Hussmann, Lenail, Guna, Mascibroda, Wagner, Adelman, Lithwick-Yanai, et~al.]{replogle2022mapping}
J.~M. Replogle, R.~A. Saunders, A.~N. Pogson, J.~A. Hussmann, A.~Lenail, A.~Guna, L.~Mascibroda, E.~J. Wagner, K.~Adelman, G.~Lithwick-Yanai, et~al.
\newblock Mapping information-rich genotype-phenotype landscapes with genome-scale perturb-seq.
\newblock \emph{Cell}, 185\penalty0 (14):\penalty0 2559--2575, 2022.

\bibitem[Roohani et~al.(2024{\natexlab{a}})Roohani, Huang, and Leskovec]{Roohani2024}
Y.~Roohani, K.~Huang, and J.~Leskovec.
\newblock Predicting transcriptional outcomes of novel multigene perturbations with gears.
\newblock \emph{Nature Biotechnology}, 42\penalty0 (6):\penalty0 927--935, 2024{\natexlab{a}}.
\newblock \doi{10.1038/s41587-023-01905-6}.
\newblock URL \url{https://doi.org/10.1038/s41587-023-01905-6}.

\bibitem[Roohani et~al.(2024{\natexlab{b}})Roohani, Huang, and Leskovec]{roohani2024predicting}
Y.~Roohani, K.~Huang, and J.~Leskovec.
\newblock Predicting transcriptional outcomes of novel multigene perturbations with gears.
\newblock \emph{Nature Biotechnology}, 42\penalty0 (6):\penalty0 927--935, 2024{\natexlab{b}}.

\bibitem[Schulman et~al.(2017)Schulman, Wolski, Dhariwal, Radford, and Klimov]{schulman2017proximal}
J.~Schulman, F.~Wolski, P.~Dhariwal, A.~Radford, and O.~Klimov.
\newblock Proximal policy optimization algorithms.
\newblock \emph{arXiv preprint arXiv:1707.06347}, 2017.

\bibitem[Shao et~al.(2024)Shao, Wang, Zhu, Xu, Song, Bi, Zhang, Zhang, Li, Wu, et~al.]{shao2024deepseekmath}
Z.~Shao, P.~Wang, Q.~Zhu, R.~Xu, J.~Song, X.~Bi, H.~Zhang, M.~Zhang, Y.~Li, Y.~Wu, et~al.
\newblock Deepseekmath: Pushing the limits of mathematical reasoning in open language models.
\newblock \emph{arXiv preprint arXiv:2402.03300}, 2024.

\bibitem[Sza{\l}ata et~al.(2024)Sza{\l}ata, Benz, Cannoodt, Cortes, Fong, Kuppasani, Lieberman, Liu, Mas-Rosario, Meinl, et~al.]{szalata2024benchmark}
A.~Sza{\l}ata, A.~Benz, R.~Cannoodt, M.~Cortes, J.~Fong, S.~Kuppasani, R.~Lieberman, T.~Liu, J.~A. Mas-Rosario, R.~Meinl, et~al.
\newblock A benchmark for prediction of transcriptomic responses to chemical perturbations across cell types.
\newblock \emph{Advances in Neural Information Processing Systems}, 37:\penalty0 20566--20616, 2024.

\bibitem[Szklarczyk et~al.(2023)Szklarczyk, Kirsch, Koutrouli, Nastou, Mehryary, Hachilif, Gable, Fang, Doncheva, Pyysalo, et~al.]{szklarczyk2023string}
D.~Szklarczyk, R.~Kirsch, M.~Koutrouli, K.~Nastou, F.~Mehryary, R.~Hachilif, A.~L. Gable, T.~Fang, N.~T. Doncheva, S.~Pyysalo, et~al.
\newblock The string database in 2023: protein--protein association networks and functional enrichment analyses for any sequenced genome of interest.
\newblock \emph{Nucleic acids research}, 51\penalty0 (D1):\penalty0 D638--D646, 2023.

\bibitem[Tang et~al.(2025)Tang, Yu, Chen, Cui, Shao, Wang, Wu, Zhuang, Shi, Huang, et~al.]{tang2025cellforge}
X.~Tang, Z.~Yu, J.~Chen, Y.~Cui, D.~Shao, W.~Wang, F.~Wu, Y.~Zhuang, W.~Shi, Z.~Huang, et~al.
\newblock Cellforge: agentic design of virtual cell models.
\newblock \emph{arXiv preprint arXiv:2508.02276}, 2025.

\bibitem[Wang et~al.(2024)Wang, Liu, Zhao, Cheng, and Zhao]{wang2024modeling}
G.~Wang, T.~Liu, J.~Zhao, Y.~Cheng, and H.~Zhao.
\newblock Modeling and predicting single-cell multi-gene perturbation responses with sclambda.
\newblock \emph{bioRxiv: the preprint server for biology}, 2024.

\bibitem[Weber(2012)]{Weber2012}
M.~Weber.
\newblock Experiment in biology.
\newblock In E.~N. Zalta, editor, \emph{The Stanford Encyclopedia of Philosophy}. Metaphysics Research Lab, Stanford University, 2012.
\newblock Spring 2012 Edition.

\bibitem[Wenkel et~al.(2025)Wenkel, Tu, Masschelein, Shirzad, Eastwood, Whitfield, Bendidi, Russell, Hodgson, Mesbahi, et~al.]{wenkel2025txpert}
F.~Wenkel, W.~Tu, C.~Masschelein, H.~Shirzad, C.~Eastwood, S.~T. Whitfield, I.~Bendidi, C.~Russell, L.~Hodgson, Y.~E. Mesbahi, et~al.
\newblock Txpert: Leveraging biochemical relationships for out-of-distribution transcriptomic perturbation prediction.
\newblock \emph{arXiv preprint arXiv:2505.14919}, 2025.

\bibitem[Wu et~al.()Wu, Littman, Levine, Qiu, Biancalani, Richmond, and Huetter]{wucontextualizing}
M.~Wu, R.~Littman, J.~Levine, L.~Qiu, T.~Biancalani, D.~Richmond, and J.-C. Huetter.
\newblock Contextualizing biological perturbation experiments through language.
\newblock In \emph{The Thirteenth International Conference on Learning Representations}.

\bibitem[Wu et~al.(2025)Wu, Littman, Levine, Qiu, Biancalani, Richmond, and Huetter]{wu2025contextualizing}
M.~Wu, R.~Littman, J.~Levine, L.~Qiu, T.~Biancalani, D.~Richmond, and J.-C. Huetter.
\newblock Contextualizing biological perturbation experiments through language.
\newblock \emph{arXiv preprint arXiv:2502.21290}, 2025.

\bibitem[Wu et~al.(2024)Wu, Wershof, Schmon, Nassar, Osi{\'n}ski, Eksi, Yan, Stark, Zhang, and Graepel]{wu2024perturbench}
Y.~Wu, E.~Wershof, S.~M. Schmon, M.~Nassar, B.~Osi{\'n}ski, R.~Eksi, Z.~Yan, R.~Stark, K.~Zhang, and T.~Graepel.
\newblock Perturbench: Benchmarking machine learning models for cellular perturbation analysis.
\newblock \emph{arXiv preprint arXiv:2408.10609}, 2024.

\bibitem[Zheng et~al.(2025)Zheng, Liu, Li, Chen, Yu, Gao, Dang, Liu, Men, Yang, et~al.]{zheng2025group}
C.~Zheng, S.~Liu, M.~Li, X.-H. Chen, B.~Yu, C.~Gao, K.~Dang, Y.~Liu, R.~Men, A.~Yang, et~al.
\newblock Group sequence policy optimization.
\newblock \emph{arXiv preprint arXiv:2507.18071}, 2025.

\bibitem[Zuheng et~al.(2024)Zuheng, Xu, Jain, Denton, Whitfield, Didolkar, Earnshaw, and Hartford]{zuheng2024automateddiscoverypairwiseinteractions}
Zuheng, Xu, M.~Jain, A.~Denton, S.~Whitfield, A.~Didolkar, B.~Earnshaw, and J.~Hartford.
\newblock Automated discovery of pairwise interactions from unstructured data, 2024.
\newblock URL \url{https://arxiv.org/abs/2409.07594}.

\end{thebibliography}
\bibliographystyle{abbrvnat}

\newpage

\newpage

\appendix
\onecolumn

\begin{table*}[t]
\centering
\scriptsize 
\setlength{\tabcolsep}{2.5pt} 

\newcommand{\res}[2]{$#1_{\pm #2}$}
\newcommand{\bres}[2]{\textbf{#1} \tiny(\textbf{#2})}

\resizebox{\textwidth}{!}{
\begin{tabular}{l cc cc cc cc}
\toprule
& \multicolumn{2}{c}{\textbf{K562}} & \multicolumn{2}{c}{\textbf{RPE1}} & \multicolumn{2}{c}{\textbf{HEPG2}} & \multicolumn{2}{c}{\textbf{Jurkat}} \\
\cmidrule(lr){2-3} \cmidrule(lr){4-5} \cmidrule(lr){6-7} \cmidrule(lr){8-9}
\textbf{Model} & Pear. $\Delta (\uparrow)$ & Retr. $(\uparrow)$ & Pear. $\Delta (\uparrow)$ & Retr. $(\uparrow)$ & Pear. $\Delta (\uparrow)$ & Retr. $(\uparrow)$ & Pear. $\Delta (\uparrow)$ & Retr. $(\uparrow)$ \\ 
\midrule
LLM, RL + KG on 4 Cell lines & \res{0.617}{.014} & \res{0.840}{.013} & \res{0.671}{.014} & \res{0.643}{.017} & \res{0.485}{.016} & \res{0.655}{.017} & \res{0.504}{.014} & \res{0.734}{.016} \\
LLM, RL + KG on 3 Cell lines & \res{0.608}{.015} & \res{0.834}{.013} & \res{0.666}{.014} & \res{0.639}{.017} & \res{0.479}{.016} & \res{0.649}{.017} & \res{0.496}{.014} & \res{0.729}{.016} \\
\midrule
KG                           & \res{0.597}{.015} & \res{0.817}{.014} & \res{0.662}{.014} & \res{0.629}{.017} & \res{0.478}{.016} & \res{0.638}{.017} & \res{0.490}{.014} & \res{0.718}{.016} \\ 
\bottomrule
\end{tabular}
}
\caption{Model performance across all four cell lines when only training on 3 and missing out K562 vs training on all 4. Scores are reported as Mean$_{\pm \text{se}}$ for standard errors. All metrics are calculated relative to the predicted and true deltas.}
\label{tab:model_performance_flat_modified}
\end{table*}

\section{Experiment Details}\label{ap:experiment_details}

\subsection{Metric Definitions}\label{ap:metric_definitions}

\begin{align}
    r(\hat{\delta}_p,{\delta}_p) = \frac{\sum\circbrac{\hat{\delta}_{p,i} - \mu({\hat{\delta}_{p}})}  \Big( {\bar{\delta}_{p,i} - \mu({\bar{\delta}_{p}}) \Big) }  }{\sqrt{\sum\circbrac{\hat{\delta}_{p,i} - \mu({\hat{\delta}_{p}})}^2 \sum \Big( {\bar{\delta}_{p,i} - \mu({\bar{\delta}_{p}}) \Big)^2 } }},
\end{align}
where  $\mu(\cdot)$ is the mean of the vector over its entries. We extend this to matrices by computing it in a row wise fashion, so $r\circbrac{\hat{\Delta},\bar{\Delta}}$ is a matrix whose $i,j$'th entry is $r\circbrac{\hat{\delta}_i,\bar{\delta}_j}$.  as follows:
\begin{align}
    d_{\mathrm{Rank},m,p}(\hat{\delta}_p,\bar{\Delta}) &= \frac{1}{\norm{P_{\mathrm{Te}}} -1} \sum_{q\in P_{\mathrm{Te}}\setminus\setbrac{p}} \mathbbm{1}\setbrac{m(\hat{{\delta}_p},\bar{\delta}_p) \geq m(\hat{{\delta}_p},\bar{\delta}_q)} \\ 
    d_{\mathrm{Retrieval},m}(\hat{\Delta},\bar{\Delta}) &= \frac{1}{\norm{P_{\mathrm{Te}}}} \sum_{p\in P_{\mathrm{Te}}} d_{\mathrm{Rank},m,p}(\hat{\delta}_p,\bar{\Delta})
\end{align}
The retrieval metric is between $0$ and $1$ with higher scores better if the original metric $m$ prefers higher scores and lower better otherwise. A score of $0.5$ would be attained by a random or constant predictor. In terms of the matrix, $r\circbrac{\hat{\Delta},\bar{\Delta}}$, this corresponds to the average over rows of the proportion of entries in that row that are less than or equal to the row's corresponding diagonal entry. 
\subsection{Scaled Experimental Reproducibility}\label{ap:scaled_exp_repro}

To establish a theoretical upper bound for model performance, we compute the experimental reproducibility of the perturbation data. The standard approach involves partitioning the set of cells observed for a specific perturbation $p$ into two disjoint subsets, $\mathcal{D}_{p}^{(1)}$ and $\mathcal{D}_{p}^{(2)}$, each containing approximately $N/2$ cells. We then compute the mean expression deltas $\bar{\delta}^{(1)}$ and $\bar{\delta}^{(2)}$ for each subset and calculate the metric of interest (here, Pearson correlation) between them.

However, this split-half approach yields an underestimate of the true experimental reproducibility of the full dataset used for evaluation. According to the Central Limit Theorem, the standard error of the mean scales with $1/\sqrt{N}$. By the delta method, the Pearson $\Delta$ will therefore also converge to the true value at rate $1/\sqrt{N}$. This justifies scaling the experimental reproducibility estimates by $1/\sqrt{2}$ to get a sample corrected estimate of the experimental reproducibility as:
\begin{equation}
    \Delta_{\text{scaled}} = 1 - \frac{1}{\sqrt{2}} \left( 1 - \Delta\left(\bar{\delta}^{(1)}, \bar{\delta}^{(2)}\right) \right)
\end{equation}
which we use throughout.

\section{Graph Neighbour Analysis Across Cell Lines}\label{ap:graph_across_cell}

\subsection{Cross Cell Line plots}
Here we present versions of the plots in Section \ref{sec:empirical} across all cell lines.

\begin{figure*}[htbp]
    \centering
    \includegraphics[width=\textwidth]{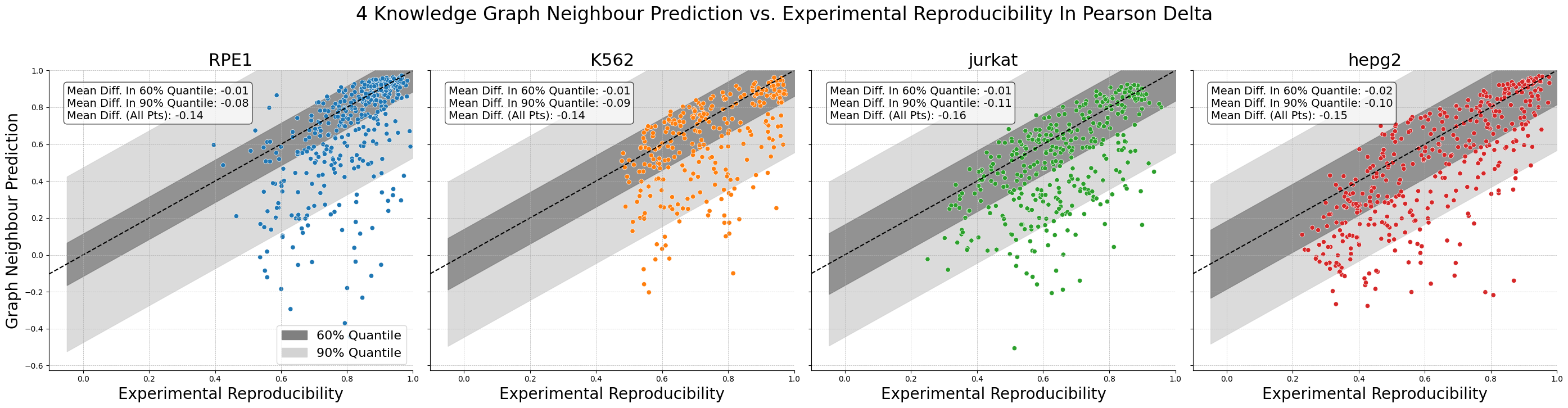}
    \caption{Plots of the by gene Pearson delta for the experimental reproducibility vs the graph neighbour with the 4 knowledge graphs from \citet{wenkel2025txpert}. Results are for all 4 cell lines from \cite{replogle2022mapping}. We add the line $y=x$ to each plot where the errors of an optimal predictor would be normally distributed around this line. To further understand the performance gap we also plot $60$ and $90$ percent quantiles for the absolute difference between the $x$ and $y$ values, and the mean of the difference in each quantile is added to the legend. This shows that for 60\% of genes the graph neighbour prediction is almost optimal and that there is a long tail in performance drop off, with the 60-90\% genes corresponding to roughly two thirds of the performance gap and the final 10\% of genes responsible for the rest. }
    \label{fig:exp_repro_vs_neighbour}
\end{figure*}

\begin{figure}[H]
    \centering
    \includegraphics[width=0.5\linewidth]{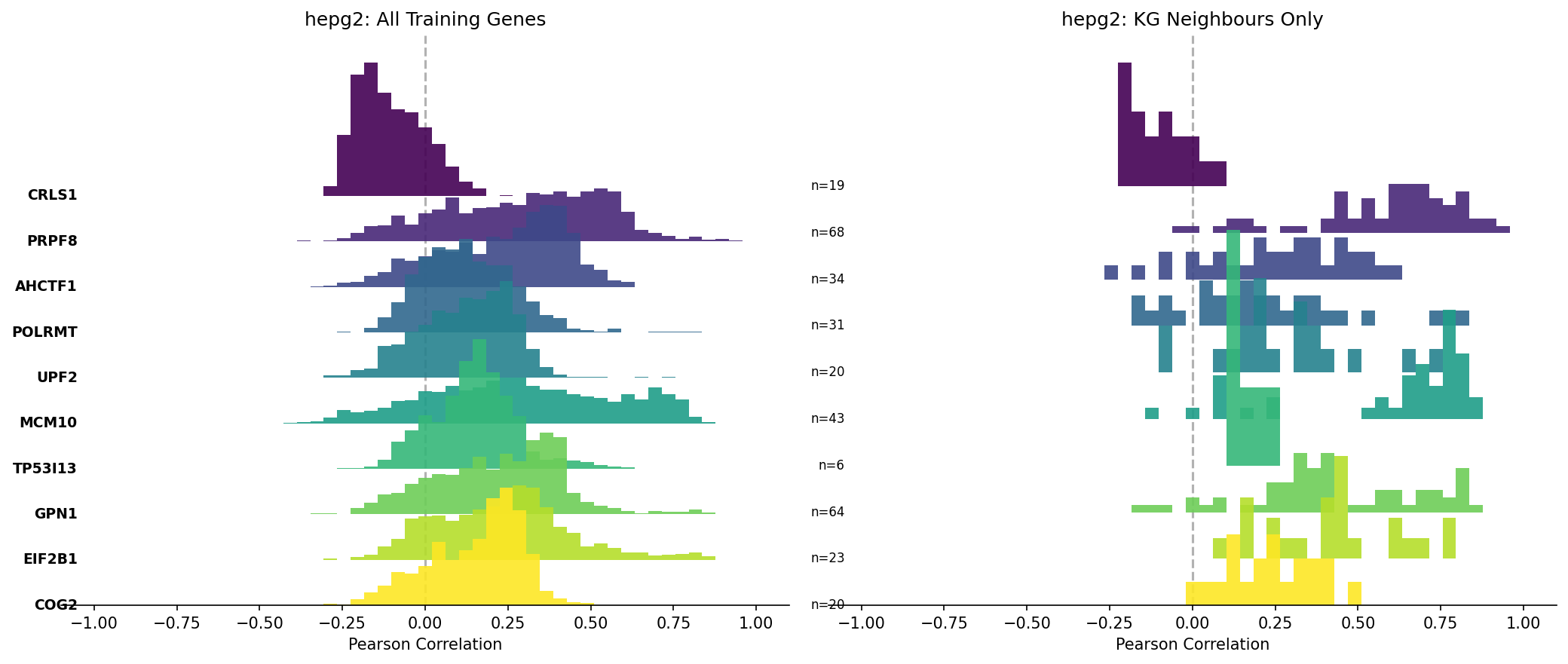}
    \caption{Ridge plots for the \textbf{HepG2} cell line comparing the distribution of Pearson correlations between target genes (y-axis) and potential predictor genes (x-axis). The left column (``All Training Genes'') displays the background correlation distribution against all genes in the training set; the right column (``KG Neighbours Only'') displays the distribution restricted to genes identified as neighbours in the Knowledge Graph.}
    \label{fig:hepg2}
\end{figure}

\begin{figure}[H]
    \centering
    \includegraphics[width=0.5\linewidth]{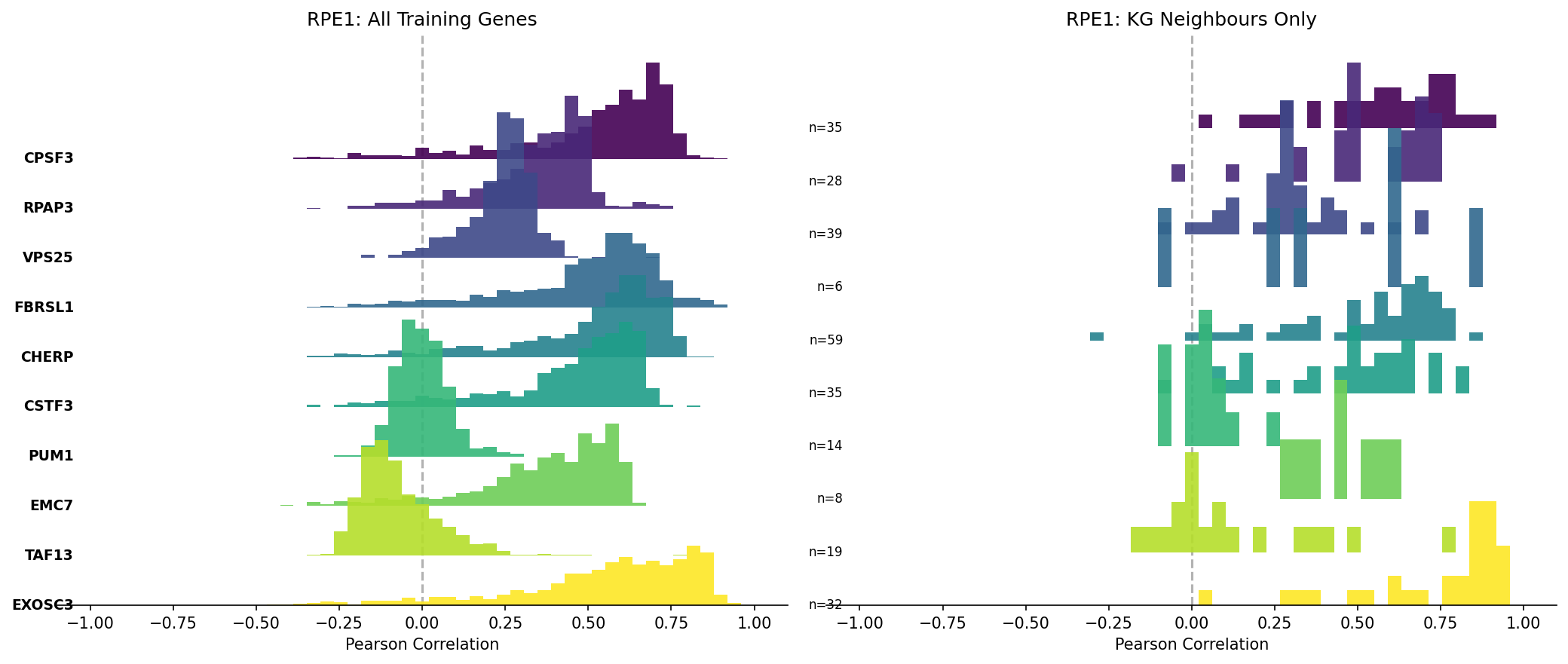}
    \caption{Ridge plots for the \textbf{RPE1} cell line comparing Pearson correlation distributions. As in Figure \ref{fig:hepg2}, the left column (``All Training Genes'') shows the background distribution for target genes, while the right column (``KG Neighbours Only'') highlights the distribution for specific Knowledge Graph neighbours.}
    \label{fig:rpe1}
\end{figure}

\begin{figure}[H]
    \centering
    \includegraphics[width=0.5\linewidth]{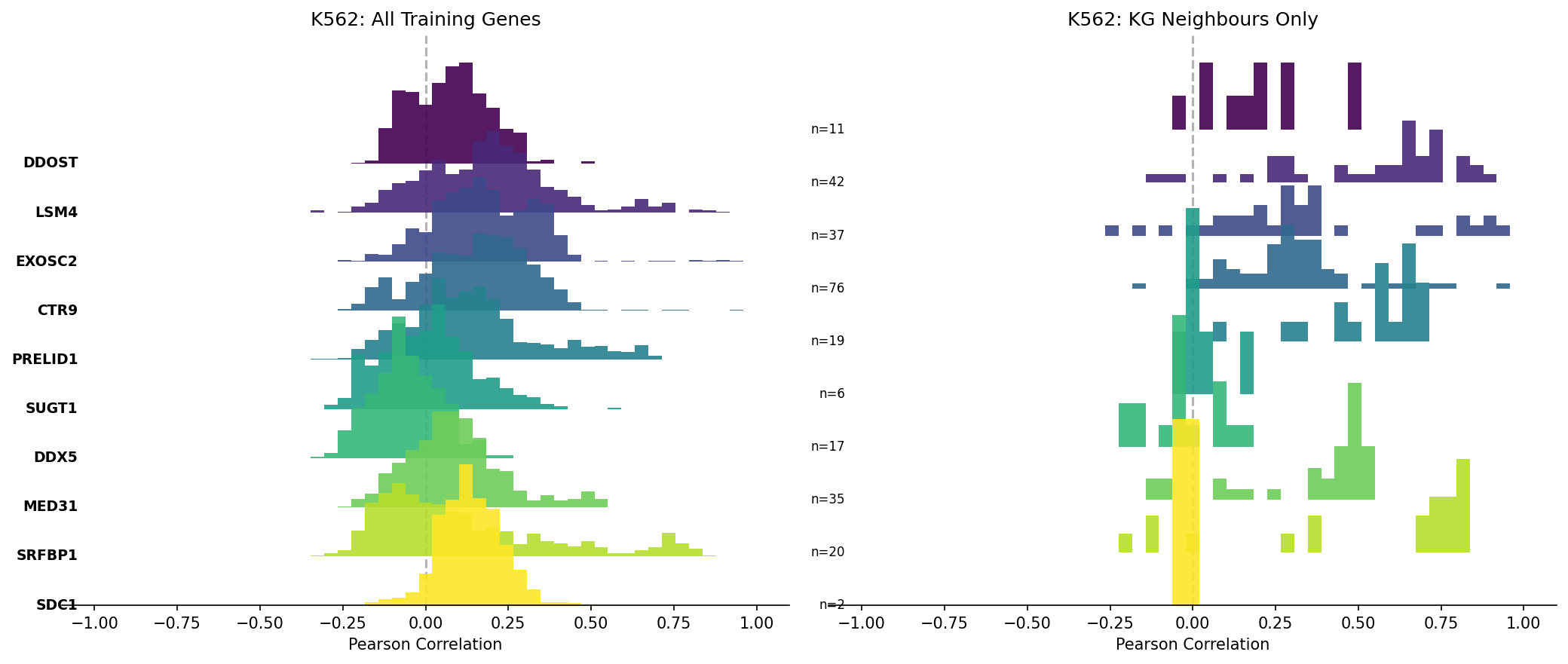}
    \caption{Ridge plots for the \textbf{K562} cell line comparing the distribution of Pearson correlations. The target-predictor relationship is shown for the full training set (left) versus restricted Knowledge Graph neighbours (right), demonstrating the correlation density shift for this specific cell line.}
    \label{fig:k562}
\end{figure}

\begin{figure}[H]
    \centering
    \includegraphics[width=0.5\linewidth]{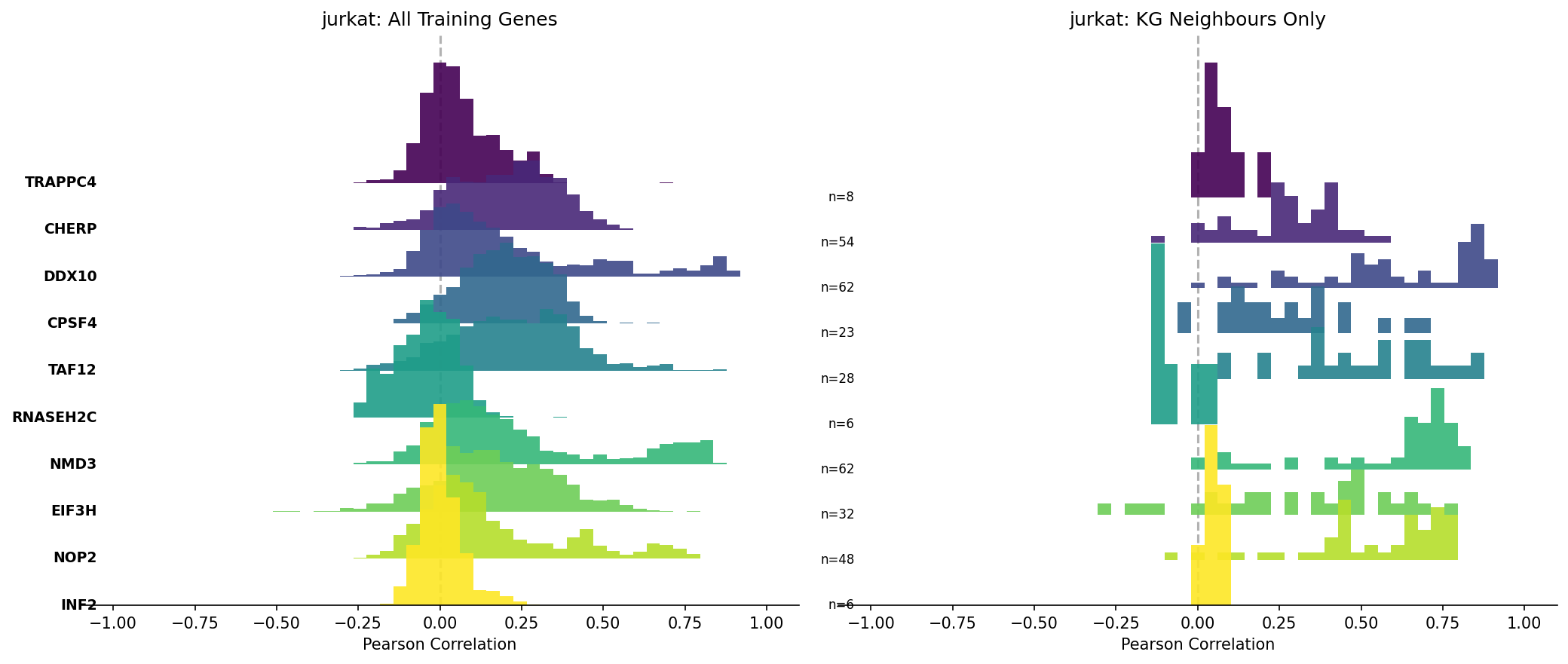}
    \caption{Ridge plots for the \textbf{Jurkat} cell line comparing Pearson correlation distributions between target genes and potential predictor genes. The data illustrates the contrast between the total training gene background (left) and the targeted Knowledge Graph neighbour distribution (right).}
    \label{fig:jurkat}
\end{figure}

\begin{figure}[H]
    \centering
    \includegraphics[width=0.7\linewidth]{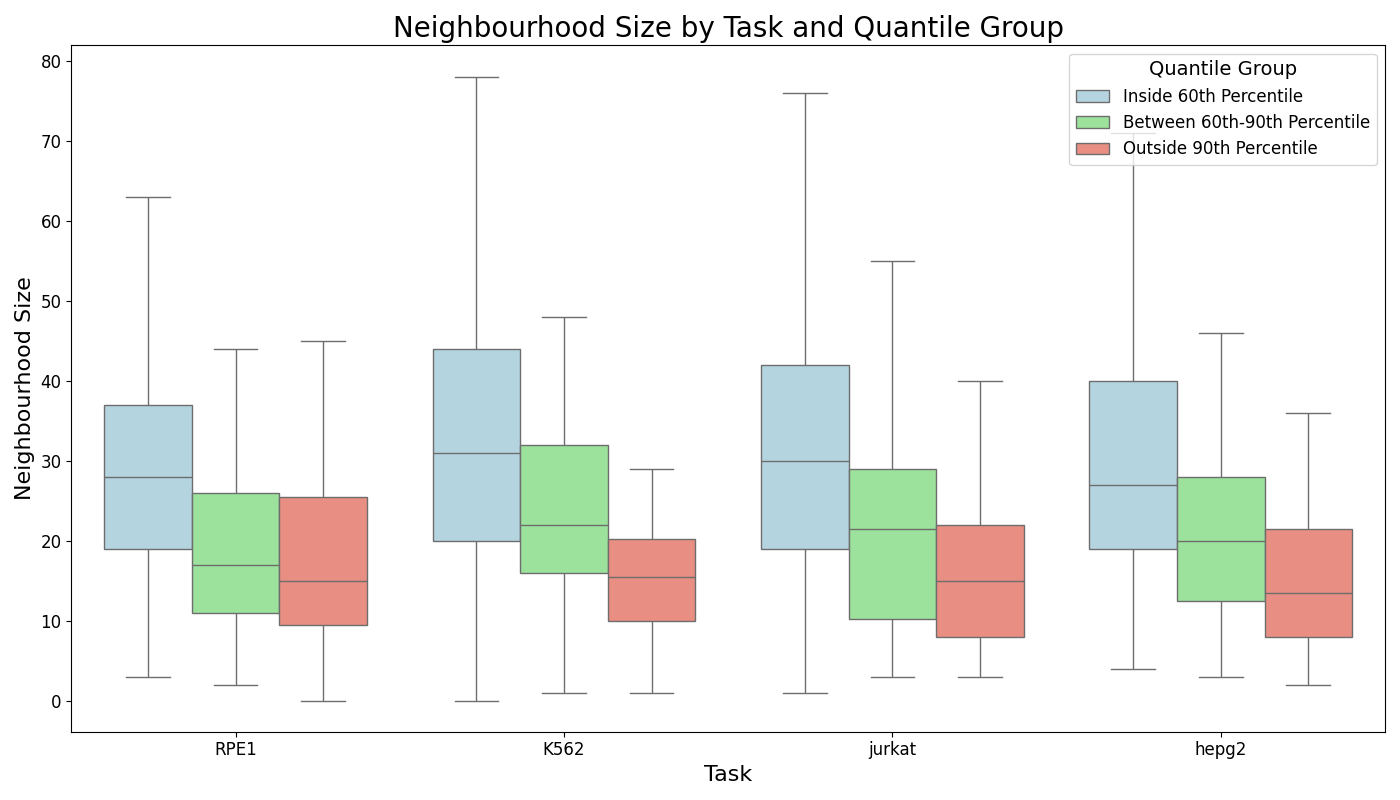} 
    \caption{Knowledge Graph neighbourhood size versus predictive performance. Genes are stratified into three groups based on the error of the graph neighbour predictor: best performing (Inside 60th Percentile), middle (60th-90th Percentile), and worst performing (Outside 90th Percentile). The box plots demonstrate that genes which are difficult to predict (red) possess significantly fewer neighbours in the knowledge graph across all four cell lines, suggesting that poor performance is driven by a lack of available signal in the biological graph.}
    \label{fig:neighbourhood_size}
\end{figure}
\subsection{Predictive Performance vs Delta Magnitude and Control Corelation}\label{ap:performance_corr}
In this Appendix we present additional results showing the correlation between the predictive performance of the graph neighbour baseline vs the delta magnitude of the perturbation and how control like a perturbation looks, we present this in terms of Pearson delta, MSE, and MAE. As reported in the main text, in Pearson delta more control like perturbations are harder predict, which is consistent with them being noisier.

Interestingly, we find this trend is reversed for mean square error and mean absolute error. For these metrics, less control like perturbations with higher magnitude deltas are harder to predict for the graph neighbour predictor. We attribute this to the fact that the predictive performance is driven by getting the magnitude of the perturbation correct, instead of the direction vs control. \citet{liu2025effects} argues for metrics based upon direction such as cosine similarity and Pearson delta---which we report---for this reason. 
\begin{figure}[H]
    \centering
    \begin{subfigure}[b]{1.0\linewidth}
        \centering
        \includegraphics[width=0.9\linewidth]{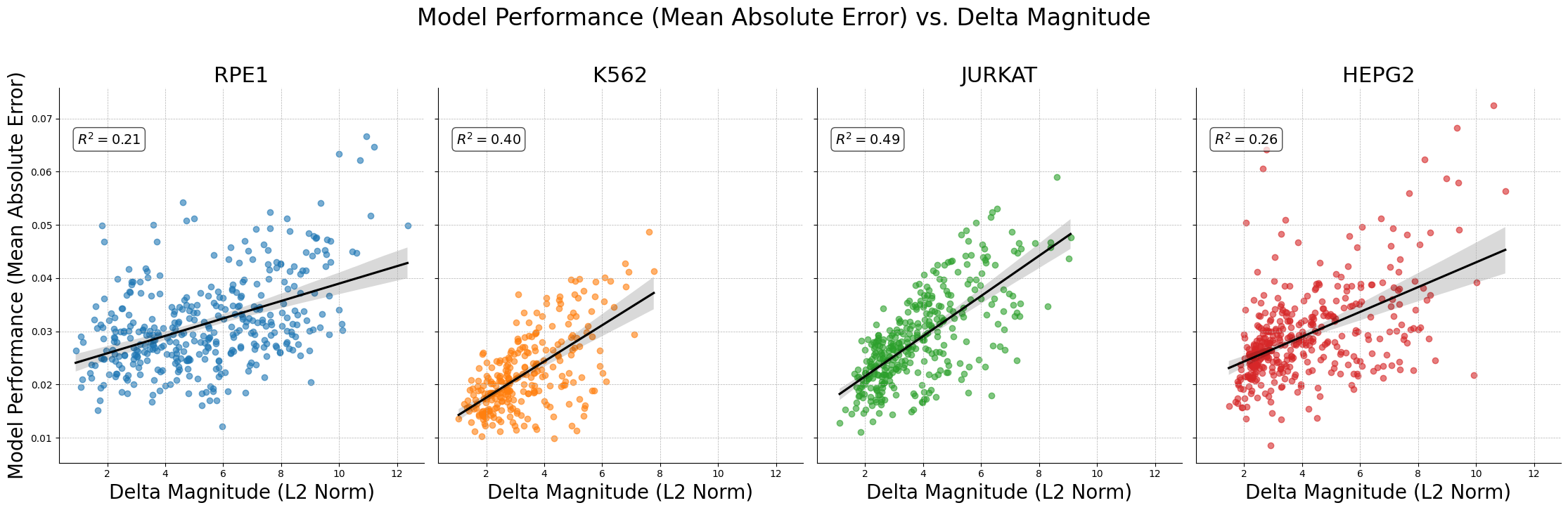}
        \caption{Mean Absolute Error (MAE)}
        \label{fig:mae_mag}
    \end{subfigure}
    
    \vspace{0.5cm} 

    \begin{subfigure}[b]{1.0\linewidth}
        \centering
        \includegraphics[width=0.9\linewidth]{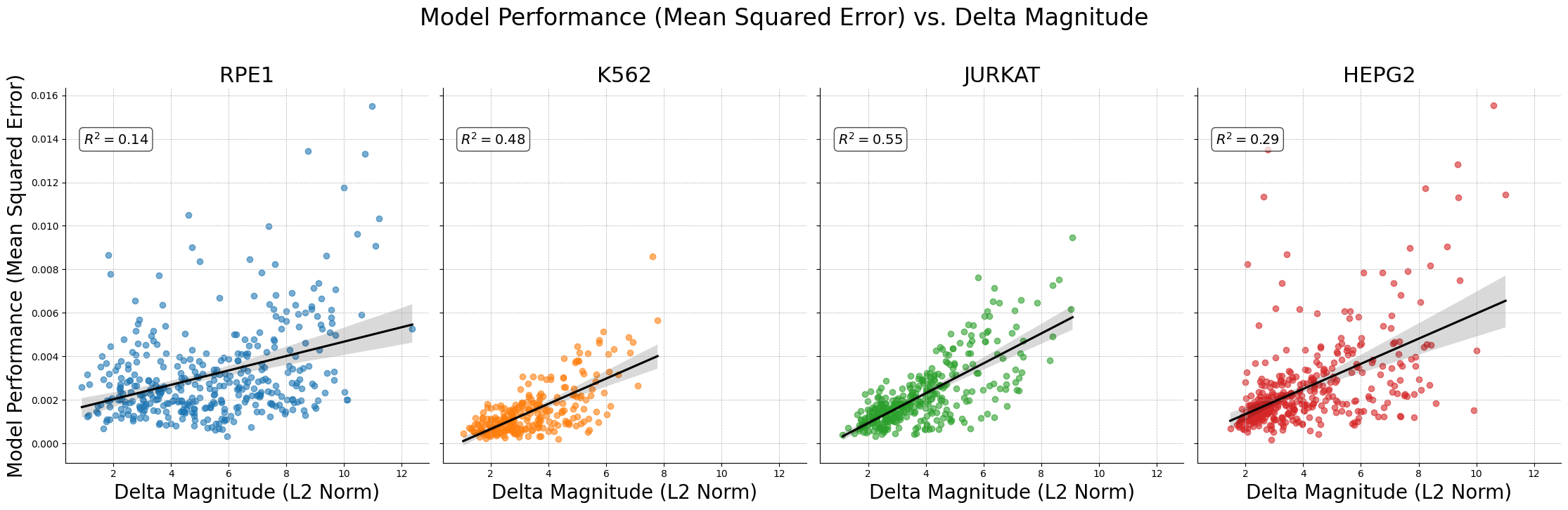}
        \caption{Mean Squared Error (MSE)}
        \label{fig:mse_mag}
    \end{subfigure}

    \vspace{0.5cm} 

    \begin{subfigure}[b]{1.0\linewidth}
        \centering
        \includegraphics[width=0.9\linewidth]{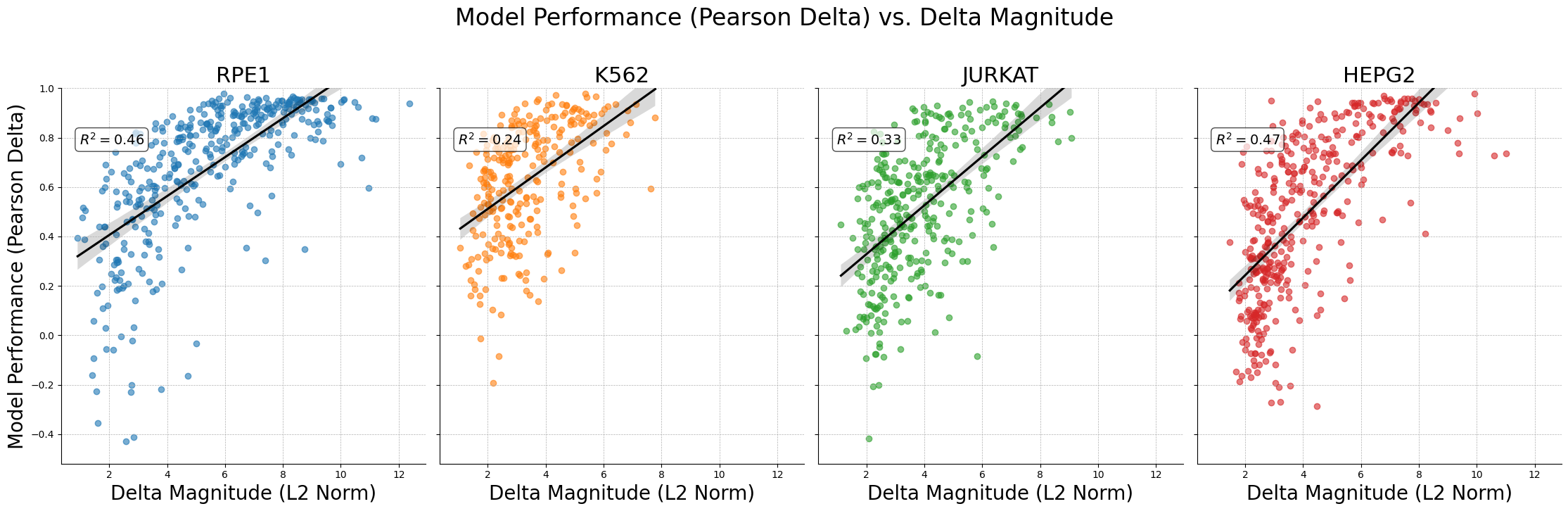}
        \caption{Pearson Delta}
        \label{fig:pearson_mag}
    \end{subfigure}

    \caption{Predictive performance versus perturbation magnitude. The metrics display diverging trends relative to magnitude: samples with larger magnitudes are easier to predict according to Pearson delta (showing a positive correlation), whereas they appear harder to predict when measured by MSE and MAE (showing increased error with magnitude).}
    \label{fig:performance_vs_magnitude}
\end{figure}

\begin{figure}[H]
    \centering
    \begin{subfigure}[b]{1.0\linewidth}
        \centering
        \includegraphics[width=0.9\linewidth]{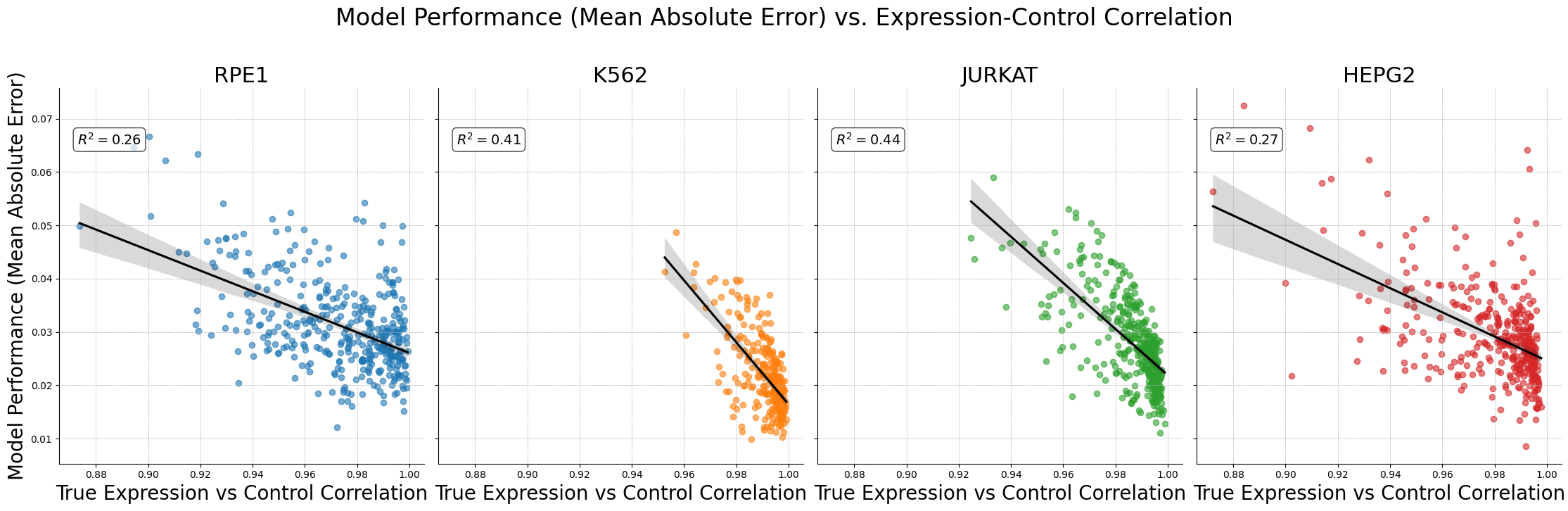}
        \caption{Mean Absolute Error (MAE)}
        \label{fig:mae_corr}
    \end{subfigure}
    
    \vspace{0.5cm} 

    \begin{subfigure}[b]{1.0\linewidth}
        \centering
        \includegraphics[width=0.9\linewidth]{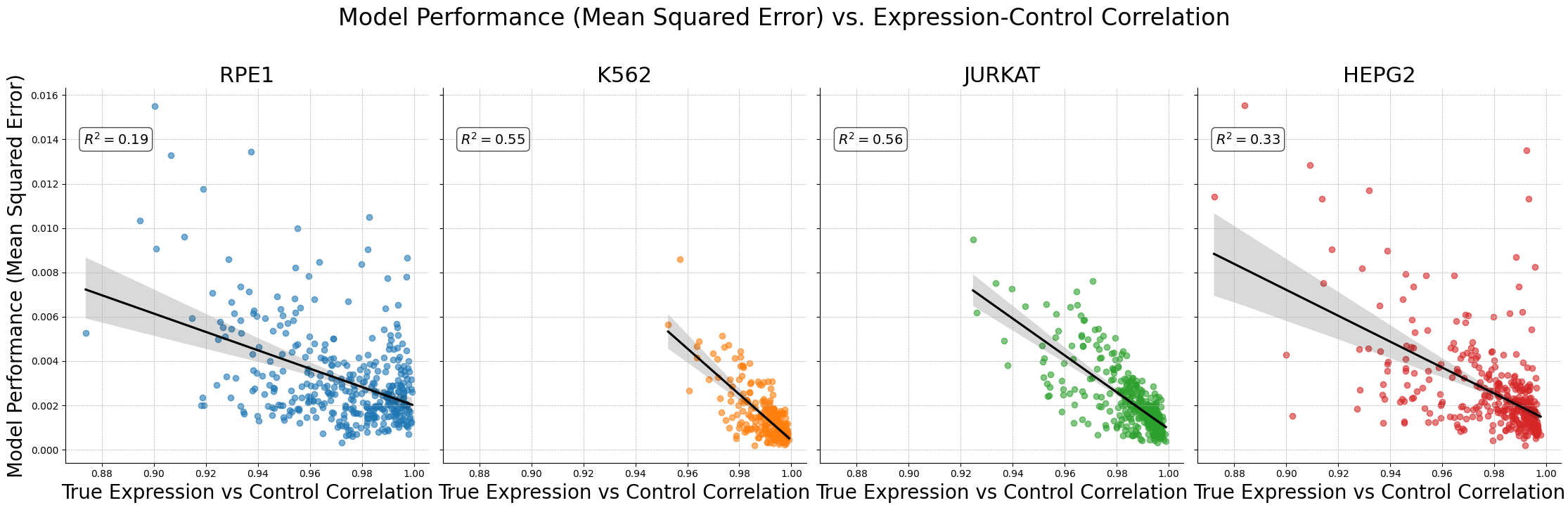}
        \caption{Mean Squared Error (MSE)}
        \label{fig:mse_corr}
    \end{subfigure}

    \vspace{0.5cm} 

    \begin{subfigure}[b]{1.0\linewidth}
        \centering
        \includegraphics[width=0.9\linewidth]{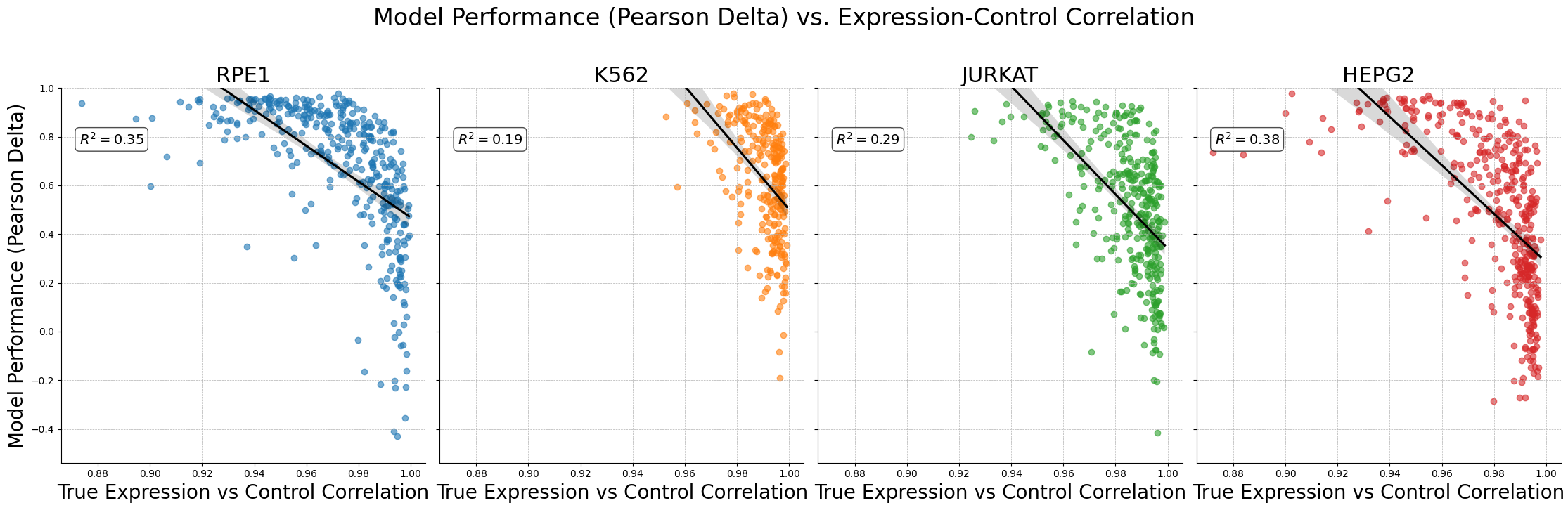}
        \caption{Pearson Delta}
        \label{fig:pearson_corr}
    \end{subfigure}

    \caption{Predictive performance versus expression-control correlation. While the Pearson delta metric exhibits a strong positive correlation with cells looking more control-like, the relationship is inverted for Mean Squared Error (MSE) and Mean Absolute Error (MAE), where lower error corresponds to higher similarity to controls.}
    \label{fig:performance_vs_control_corr}
\end{figure}

\section{LangPert Vs Graph Neighbour Predictor}\label{ap:langpert_vs_kg}

Here we present a brief analysis of the overlap in performance between LangPert and the graph neighbour approach. Firstly, we find that on average 50\% of the genes proposed by the LangPert with Gemini 2.5 pro lie in the union of all four knowledge graphs, and 30\% lie in the set for Claude 4 Sonnet. Moreover, as the following plot shows, subsetting down to the genes which lie in the intersection does not lead to a discernable decrease in performance.

\begin{figure}[H]
    \centering
    \includegraphics[width=0.4\linewidth]{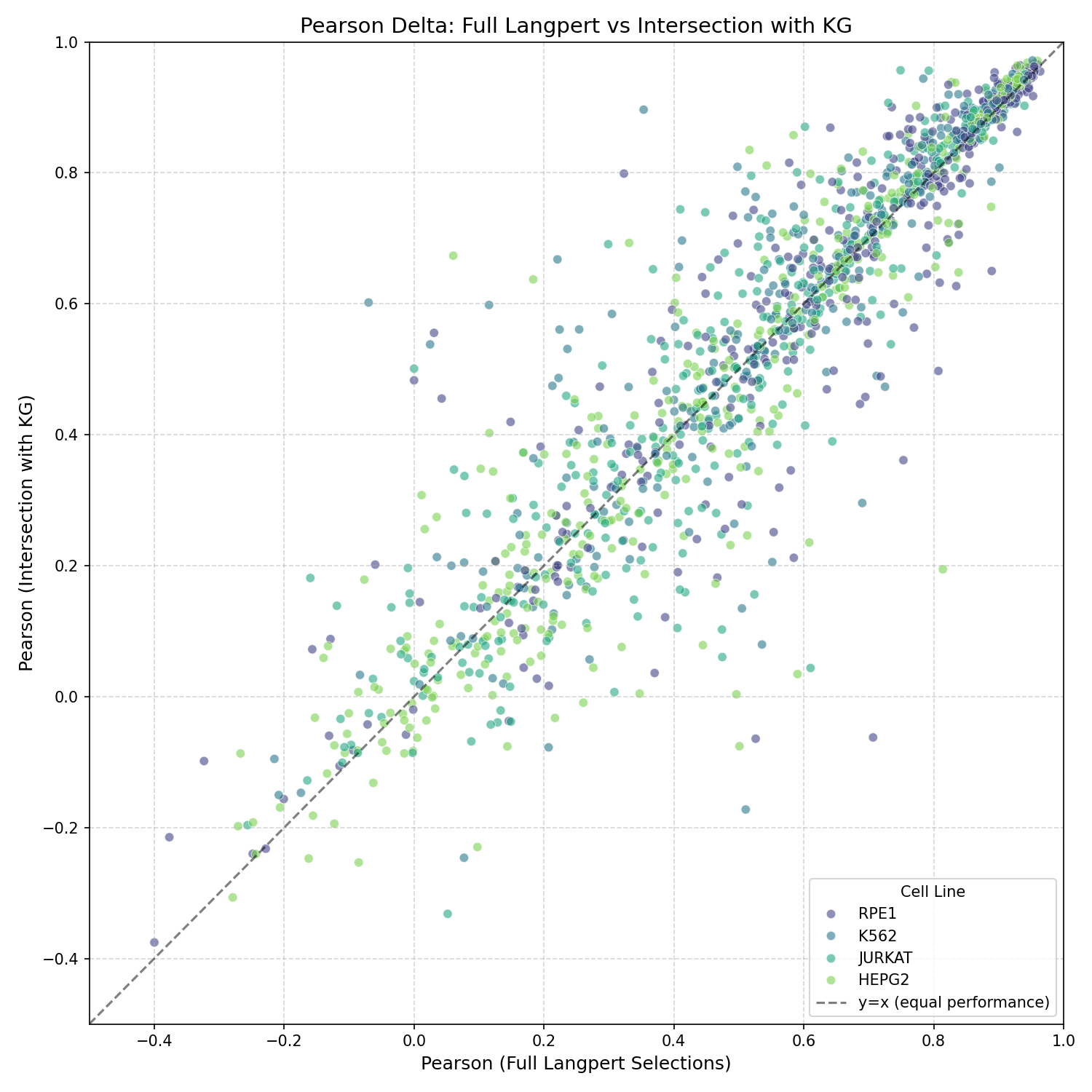}
    \label{fig:placeholder}
\end{figure}
\section{Model Details}\label{ap:model_details}

\subsection{LangPert style Approach}

For LangPert, we evaluate using Gemini 2.5 Pro and Claude 4 Sonnet, asking each to give on set of responses per answer. Our prompts are given as follows:

\noindent\begin{minipage}{\linewidth}
    
    \begin{tcolorbox}[colback=gray!10, colframe=black!75, title=\textbf{System Prompt}, arc=2mm, sharp corners=south, bottomrule=0pt]
    You are an expert computational biologist specializing in functional genomics and pathway analysis. Provide concise, evidence-based reasoning and strictly adhere to all formatting instructions.
    \end{tcolorbox}%
    \nointerlineskip
    \begin{tcolorbox}[colback=white, colframe=black!75, title=\textbf{User Prompt}, arc=2mm, sharp corners=north, toprule=0pt]
    Your goal is to choose genes from the provided gene list that will produce a similar average cellular response to a knockout of \texttt{\{gene\}} in the \texttt{\{cell\_line\}} cell line.
    
    \vspace{0.3cm}
    
    You must select genes from the following gene list: \texttt{\{gene\_list\}}. Detail your reasoning for each gene you select and why you think it will produce a similar cellular response. Focus on identifying protein-encoding genes from the list that share a functional relationship (e.g., same pathway, protein complex, similar biological process) with \texttt{\{gene\}}.
    
    \vspace{0.3cm}

    Give your final answer as a list of selected genes in a boxed format as follows: 
    \begin{center}
    \texttt{\textbackslash boxed\{\{[gene1, gene2, gene3, ...]\}\}}
    \end{center}

    \vspace{0.2cm}
    \noindent\rule{\linewidth}{0.4pt} 
    \vspace{0.2cm}

    To help, here are some key related terms about \texttt{\{gene\}}, the gene you are trying to predict the response of (This may be empty if the gene is not in the database):
    \newline
    \texttt{\{gene\_info\}}

    \vspace{0.3cm}

    Additionally, here is some background information about the \texttt{\{cell\_line\}} cell line:
    \newline
    \texttt{\{cell\_line\_background\}}
    \end{tcolorbox}

\end{minipage}

\subsection{RL, LLM and Knowledge Graph Approach}\label{ap:rl_train}

For training the RL model, we use \citet{primeintellect2025prime-rl}, training on 4 H100s for 192 hours total. We use the default prime-rl GRPO implementation with unchanged hyper-parameters.

For the hyperparameters of our reward, we set $\lambda=5$ and $\epsilon=0.01$.
\subsubsection{Prompts}\label{ap:RL_LLLM_prompt}

\noindent\begin{minipage}{\linewidth}
    
    \begin{tcolorbox}[colback=gray!5, colframe=black!75, title=\textbf{System Prompt}, arc=2mm, sharp corners=south, breakable=false, bottomrule=0pt]
    You are an expert computational biologist specializing in functional genomics and pathway analysis. Provide concise, evidence-based reasoning and strictly adhere to all formatting instructions.
    \end{tcolorbox}
    
    \begin{tcolorbox}[colback=white, colframe=black!75, title=\textbf{User Prompt}, arc=2mm, sharp corners=north, breakable=false, toprule=0pt]
    Your goal is to choose genes that will produce a similar average cellular response to a knockout of \texttt{\{gene\}} in the \texttt{\{cell\_line\}} cell line. To start off, you are given the immediate neighbours of this gene in string db: \texttt{\{neighbourhood\}}. 

    \vspace{0.3cm}

    You can also choose to add or remove genes from the neighbourhood list, where the genes you add must come from the following gene list: \texttt{\{gene\_list\}}. Detail your reasoning for each gene you add or remove and why you think it will produce a similar or disimilar response. Briefly detail your reasoning. Focus on identifying protein-encoding genes from the list that share a functional relationship (e.g., same pathway, protein complex) with \texttt{\{gene\}}.

    \vspace{0.3cm}

    Give your final answer of genes to add or remove in a boxed format as follows: 
    \begin{center}
    \texttt{\textbackslash boxed\{\{genes\_to\_add: [gene1, gene2, ...], genes\_to\_remove: [gene3, gene4, ...]\}\}}
    \end{center}

    \vspace{0.2cm}
    \noindent\rule{\linewidth}{0.4pt} 
    \vspace{0.2cm}

    To help, here are some some key related terms about \texttt{\{gene\}}, the gene you are trying to predict the response of (This may be empty if the gene is not in the database):
    \newline
    \texttt{\{gene\_info\}}

    \vspace{0.3cm}

    Additionally, here is some background information about the \texttt{\{cell\_line\}} cell line:
    \newline
    \texttt{\{cell\_line\_background\}}
    \end{tcolorbox}

\end{minipage}

%

\author{%
  David S.~Hippocampus\thanks{Use footnote for providing further information
    about author (webpage, alternative address)---\emph{not} for acknowledging
    funding agencies.} \\
  Department of Statistics\\
  Cranberry-Lemon University\\
  Pittsburgh, PA 15213 \\
  \texttt{hippo@cs.cranberry-lemon.edu} \\
}

\end{document}